%% file: Paper.tex
\documentclass[sigplan,screen]{acmart}

\usepackage{subfig}
\usepackage{ctable}
\usepackage{booktabs}
\usepackage{longtable}
\usepackage{graphicx}
\usepackage{float}
\usepackage{multirow}
\usepackage{lscape}
\usepackage[T1]{fontenc}
\usepackage{algorithm,algorithmicx}
\usepackage{algcompatible}
\usepackage{url}
\usepackage{amsmath, amsthm, amssymb}
\usepackage{amsfonts}
\usepackage{enumitem}

\usepackage[normalem]{ulem}
\usepackage{pifont}

\usepackage{xspace}
\newcommand{\projectname}{PatDNN\xspace}

\usepackage[format=plain,indention=.6cm, font=small, labelfont=bf,belowskip=3pt,aboveskip=3pt]{caption}

\usepackage{balance}

\copyrightyear{2020}
\acmYear{2020}
\setcopyright{acmlicensed}
\acmConference[ASPLOS '20]{Proceedings of the Twenty-Fifth International Conference on Architectural Support for Programming Languages and Operating Systems}{March 16--20, 2020}{Lausanne, Switzerland}
\acmBooktitle{Proceedings of the Twenty-Fifth International Conference on Architectural Support for Programming Languages and Operating Systems (ASPLOS '20), March 16--20, 2020, Lausanne, Switzerland}
\acmPrice{15.00}
\acmDOI{10.1145/3373376.3378534}
\acmISBN{978-1-4503-7102-5/20/03}

\begin{document}

% Including title, authors, footnotes, abstract
\input{tex/header.tex}
% Introduction to this paper
\input{tex/introduction.tex}
% Background of this paper
\input{tex/background.tex}

% Overview of this paper
\input{tex/overview.tex}
\input{tex/pattern-prune.tex}
\input{tex/design.tex}
\input{tex/implementation.tex}

\input{tex/evaluation.tex}
\input{tex/conclusion.tex}
\input{tex/ack.tex}

\balance
% \bibliographystyle{ACM-Reference-Format}
% \bibliography{references}
%%% -*-BibTeX-*-
%%% Do NOT edit. File created by BibTeX with style
%%% ACM-Reference-Format-Journals [18-Jan-2012].

\end{document}

%% file: tex/header.tex
\title[PatDNN]{PatDNN: Achieving Real-Time DNN Execution on Mobile Devices with Pattern-based Weight Pruning}

\input{tex/authors.tex}

%%
%% By default, the full list of authors will be used in the page
%% headers. Often, this list is too long, and will overlap
%% other information printed in the page headers. This command allows
%% the author to define a more concise list
%% of authors' names for this purpose.
\renewcommand{\shortauthors}{Wei Niu and Xiaolong Ma, et al.}

\input{tex/abstract.tex}

\begin{CCSXML}
<ccs2012>
    <concept>
        <concept_id>10010147.10010257.10010293.10010294</concept_id>
        <concept_desc>Computing methodologies~Neural networks</concept_desc>
        <concept_significance>500</concept_significance>
    </concept>
    <concept>
        <concept_id>10011007.10011006.10011041.10011047</concept_id>
        <concept_desc>Software and its engineering~Source code generation</concept_desc>
        <concept_significance>500</concept_significance>
    </concept>
    <concept>
        <concept_id>10003120.10003138.10003139.10010905</concept_id>
        <concept_desc>Human-centered computing~Mobile computing</concept_desc>
        <concept_significance>500</concept_significance>
    </concept>
</ccs2012>
\end{CCSXML}

\ccsdesc[500]{Computing methodologies~Neural networks}
\ccsdesc[500]{Software and its engineering~Source code generation}
\ccsdesc[500]{Human-centered computing~Mobile computing}

%%
%% Keywords. The author(s) should pick words that accurately describe
%% the work being presented. Separate the keywords with commas.
\keywords{Deep Neural Network, Model Compression, Compiler Optimization, Mobile Devices}

\maketitle

%% file: tex/authors.tex
%%
%% The "author" command and its associated commands are used to define
%% the authors and their affiliations.
%% Of note is the shared affiliation of the first two authors, and the
%% "authornote" and "authornotemark" commands
%% used to denote shared contribution to the research.

% Wei Niu
\author{Wei Niu}
\affiliation{%
  \institution{College of William and Mary}
  \streetaddress{}
  \city{}
  \state{}
  \postcode{}
}
\email{wniu@email.wm.edu}

% Xiaolong Ma
\author{Xiaolong Ma}
\affiliation{%
  \institution{Northeastern University}
  \streetaddress{}
  \city{}
  \country{}
 }
\email{ma.xiaol@husky.neu.ed}

% Sheng Lin
\author{Sheng Lin}
\affiliation{%
  \institution{Northeastern University}
  \streetaddress{}
  \city{}
  \country{}
}
\email{lin.sheng@husky.neu.edu}

% Shihao Wang
\author{Shihao Wang}
\affiliation{%
  \institution{Northeastern University}
  \streetaddress{}
  \city{}
  \country{}
}
\email{wang.shih@husky.neu.edu}

% Prof. Xuehai Qian
\author{Xuehai Qian}
\affiliation{%
  \institution{University of Southern California}
  \streetaddress{}
  \city{}
  \country{}
}
\email{xuehai.qian@usc.edu}

% Prof. Xue Lin
\author{Xue Lin}
\affiliation{%
  \institution{Northeastern University}
  \streetaddress{}
  \city{}
  \country{}
}
\email{xue.lin@northeastern.edu}

% Prof. Yanzhi Wang
\author{Yanzhi Wang}
\affiliation{%
  \institution{Northeastern University}
  \streetaddress{}
  \city{}
  \country{}
}
\email{yanz.wang@northeastern.edu}

% Prof. Bin Ren
\author{Bin Ren}
\affiliation{%
  \institution{College of William and Mary}
  \streetaddress{}
  \city{}
  \country{}
}
\email{bren@cs.wm.edu}

%% file: tex/abstract.tex
\begin{abstract}

%\begin{itemize}
    % \item GPU: the relationship between speed and CHW size.(By Total load time, computation/load ratio, register numbers)
    % \item Code generation: to remove redundant load/computation and branch divergence. (By jinja2)
    % \item Pattern pruning: interpretability
    % \item Predict model: a model to show the highest speed when running a CNN on mobile device
    
With the emergence of a spectrum of high-end mobile devices, many applications that
formerly required desktop-level computation capability are being transferred to these devices. 
 However, executing Deep Neural Networks (DNNs) inference is still challenging considering the high computation and storage demands, specifically, if real-time performance with high accuracy is needed.
Weight pruning of DNNs is proposed, but existing schemes represent two extremes in the design space: non-structured pruning is fine-grained, accurate, but not hardware friendly; structured pruning is coarse-grained, hardware-efficient, but with higher accuracy loss.

In this paper, we advance the state-of-the-art by introducing a new dimension, fine-grained pruning patterns inside the coarse-grained structures, revealing a previously unknown point in the design space. With the higher accuracy enabled by fine-grained pruning patterns, the unique insight is to use the compiler to re-gain and guarantee high hardware efficiency. In other words, our method achieves the best of both worlds, and is desirable across theory/algorithm, compiler, and hardware levels. The proposed PatDNN is an end-to-end framework to efficiently execute
DNN on mobile devices with the help of a novel
 model compression technique---pattern-based pruning based on an extended ADMM solution framework---and a set of thorough architecture-aware compiler/code generation-based optimizations, i.e., filter kernel reordering, compressed weight storage, register load redundancy elimination, and parameter auto-tuning.
 Evaluation results demonstrate that PatDNN outperforms three state-of-the-art end-to-end DNN frameworks, TensorFlow Lite, TVM, and Alibaba Mobile Neural Network with speedup up to $44.5\times$, $11.4\times$, and $7.1\times$, respectively, with no accuracy compromise. Real-time inference of representative large-scale DNNs (e.g., VGG-16, ResNet-50) can be achieved using mobile devices. 
 %\todo{Place holder, to revise}

%\end{itemize}

\end{abstract}

%% file: tex/introduction.tex
\section{Introduction}

%References: https://medium.com/syncedreview/deep-learning-in-real-time-inference-acceleration-and-continuous-training-17dac9438b0b

%DeftNN: http://cccp.eecs.umich.edu/papers/parkerhh-micro17.pdf

Deep learning or deep neural networks (DNNs) have become the fundamental element and core enabler of ubiquitous artificial intelligence. 
After obtaining DNN models trained with 
a huge amount of data, 
they can be 
%After training with huge amount of data, DNN models can be 
deployed for inference, perception and control tasks in various autonomous systems and internet-of-things (IoT) applications.
Recently, along with the rapid emergence of high-end mobile devices\footnote{Modern mobile platforms become increasingly sophisticated, usually equipped with both CPUs and GPUs, e.g., Qualcomm Snapdragon 855 \cite{snapdragon855} has an octa-core Kryo 485 CPU and an Adreno 640 GPU.}, executing DNNs on mobile platforms gains popularity and is quickly becoming the mainstream~\cite{deng2019deep,lane2016deepx,lane2017squeezing,ota2017deep,zhang2019deep}
for broad applications such as sensor nodes, wireless access points, smartphones, wearable devices, video streaming, augmented reality, robotics, unmanned vehicles, smart health devices, etc.~\cite{philipp2011sensor,lane2015early,boticki2010quiet,rodgers2014recent,bhattacharya2016smart}. 

Considering the nature of these applications, 
achieving {\em real-time DNN inference} is an ideal 
but yet a very challenging goal for mobile devices
%it is ideal to achieve {\em real-time DNN inference} using mobile devices. However, it is a challenging task 
due to the limited computing resources of embedded processors. 
For example, consider VGG-16 \cite{simonyan2014very}, one of the key DNN models in transfer learning with broad application scenarios. For an embedded GPU (Adreno 640, with 16-bit floating-point for weights/intermediate results), it takes 242ms to
perform inference using TVM \cite{chen2018tvm}, and is not even supported in TensorFlow-Lite (TFLite) \cite{TensorFlow-Lite} --- these are two representative mobile-oriented, end-to-end DNN inference acceleration frameworks. It is clearly far from 
real-time execution.

To achieve the goal, it is necessary to 
consider algorithm-level innovations. 
%It is natural to turn to algorithm-level improvements for facilitating real-time inference execution.
To this end, {\em DNN model compression} techniques, including \emph{weight pruning}~\cite{han2015learning,han2015deep,guo2016dynamic,dai2017nest,mao2017exploring,wen2016learning,he2017channel} and \emph{weight/activation quantization}~\cite{leng2017extremely,park2017weighted,zhou2017incremental,lin2016fixed,wu2016quantized,rastegari2016xnor,hubara2016binarized,courbariaux2015binaryconnect,courbariaux2016binarized,gupta2015deep,hubara2017quantized}, have been proposed and 
studied intensively for model storage reduction and computation acceleration. 
Early efforts on DNN model compression~\cite{han2015learning,han2015deep,guo2016dynamic,dai2017nest,mao2017exploring,wen2016learning,he2017channel} mainly rely on iterative and heuristic methods, with limited and non-uniform model compression rates.
Recently, a systematic DNN model compression framework (ADMM-NN) has been developed using the powerful mathematical optimization tool ADMM (Alternating Direction Methods of Multipliers) \cite{boyd2011distributed,hong2016convergence,liu2018zeroth}, currently achieving the best performance (in terms of model compression rate under the same accuracy) on weight pruning \cite{zhang2018systematic,ren2019ADMMNN} and one of the best on weight quantization \cite{leng2017extremely}. 

%We also point out that 8-bit fixed point representation is already supported in current mobile devices, which is compatible with this work and can be combined for further performance enhancement. 

Despite the high compression ratio, 
there is a significant gap between algorithm-level innovations and hardware-level performance 
optimizations for DNN inference acceleration.
Specifically, the general but 
{\em non-structured} weight pruning 
(i.e., arbitrary weight can be pruned)~\cite{han2015learning,guo2016dynamic}
can seriously affect processing throughput
because the indices for the compressed 
weight representation prevent achieving
high parallelism~\cite{wen2016learning,he2017channel,mao2017exploring}.
%However, we have noticed a significant gap between algorithm-level innovations and hardware-level performance improvements, for the DNN inference acceleration problem. It is currently commonly known that the general, non-structured weight pruning (arbitrary weight can be pruned) \cite{han2015learning,guo2016dynamic} is incompatible with instruction-level parallelism \cite{wen2016learning,he2017channel,mao2017exploring}. 
While ADMM-NN achieves higher and more reliable
compression ratios, hardware implementation
obstacle due to the non-structured nature still stays the same. 
%As a result, a notable acceleration is difficult to achieve even with the help of ADMM. 
Alternatively, the {\em structured} pruning \cite{wen2016learning,he2017channel,mao2017exploring},
e.g., filter and channel pruning, can generate
more hardware-friendly models but result
in relatively higher accuracy drop.
%On the other hand, we point out that current structured pruning schemes (mainly filter and channel pruning) result in relatively significant accuracy drop. 
To achieve the real-time inference for
representative DNNs in mobile devices,
it is imperative to develop 
an end-to-end DNN acceleration framework 
that achieves {\em both high accuracy 
and high hardware efficiency}.
%, achieving real-time inference for most representative DNNs in mobile devices.

%As a result, we are in urgent need of an end-to-end DNN acceleration framework that is both highly accurate and hardware friendly, achieving real-time inference for most representative DNNs in mobile devices.

We make a key observation 
that the general non-structured pruning
and current structured pruning 
represent two extremes in the design space.
In non-structured pruning, {\em any} weight can be 
pruned, while in structured pruning, the pruning 
is done for
the {\em whole filter or channel}.
Thus, non-structured pruning is 
completely {\em fine-grained}, which achieves
high compression ratio but is not hardware or 
software optimization friendly,
while structured pruning is {\em coarse-grained},
which generates hardware-efficient regular models
with higher accuracy loss. 

In this paper, we advance the state-of-the-art
by naturally introducing a new dimension,
{\em fine-grained 
pruning patterns inside the coarse-grained
structures}, revealing a previously {\em unknown}
point in design space. 
This new dimension allows more flexible
exploration of the trade-off between 
accuracy and hardware efficiency. 
In this paradigm, the key question is 
{\em how to ``recover'' the hardware efficiency
lost due to the fine-grained patterns}.
The unique insight of our solution is 
%Our \textbf{key motivation} is 
to use {\em compiler} to seamlessly 
close the gap between hardware efficiency
of fully structured pruning and the 
pattern-based ``semi-structured'' pruning.

Specifically, we propose
{\em \projectname}, a novel end-to-end mobile DNN acceleration framework that can generate highly accurate
DNN models using pattern-based pruning methods
and guarantee execution efficiency with 
compiler optimizations.
\projectname consists of two stages: (1) \emph{pattern-based training stage}, which performs kernel pattern and connectivity pruning (termed \emph{pattern-based pruning} in general) with a pattern set generation and an extended ADMM solution framework. (2) \emph{execution code generation stage}, which converts DNN models into computational graphs and applies multiple optimizations including: a high-level and fine-grained DNN layerwise representation, filter kernel reorder, load redundancy eliminations, and automatic parameter tuning. All design optimizations are general, and applicable to both mobile CPUs and GPUs.

% In our study, we focus on the weight pruning scheme while using 16-bit (half) floating point representation for both weights and intermediate results, which is supported in mobile devices and shown to incur no accuracy loss \cite{TensorFlow-Lite,Ali-MNN,chen2018tvm} for DNNs. 
% In fact, 8-bit fixed point representation
% is also supported in current mobile devices,
% the proposed techniques
% can be also be applied to this case and
% may lead to further performance enhancement. 

%\blue{
In sum, this paper makes several major contributions:
%}

\begin{itemize}%[leftmargin=*,noitemsep,nolistsep]
    \item %\blue{
    First, it proposes a novel {\em pattern-based} DNN pruning approach that achieves the benefits of both non-structured and structured pruning while avoiding their weaknesses.
    %}
    
    \item %\blue{
    Second, it enhances the recent ADMM-NN framework~\cite{ren2019ADMMNN,ye2019progressive} with pattern selection capability to map a pattern to each kernel, and train non-zero weights.
    %}
    
    \item %\blue{
    Third, it identifies the compatibility of the proposed pattern-based pruning scheme with compiler code generation, and develop multiple novel compiler optimizations for compressed DNN execution. These optimization opportunities are enabled only by our pattern-based design, and do not exist in any prior DNN execution frameworks.
    %}
    
    \item %\blue{
    Fourth, it implements an end-to-end DNN acceleration framework {\em PatDNN} on mobile platforms, compatible with modern embedded CPU and GPU architectures, achieving real-time performance on representative DNNs without accuracy loss for the first time.
    %}
\end{itemize}

%\red{SUMMARY OF RESULTS}
We compare PatDNN with three state-of-the-art end-to-end DNN frameworks on both mobile CPU and GPU, TensorFlow Lite~\cite{TensorFlow-Lite}, TVM~\cite{chen2018tvm}, and Alibaba Mobile Neural Networks~\cite{Ali-MNN} using three widely used DNNs, VGG-16, ResNet-50, and MobileNet-V2 and two benchmark datasets, ImageNet and CIFAR-10. Our evaluation results show that PatDNN achieves up to $44.5\times$ speedup without any accuracy compromise. Using Adreno 640 embedded GPU, PatDNN achieves 18.9ms inference time of VGG-16 on ImageNet dataset. To the best of our knowledge, it is the first time to achieve real-time execution of such representative large-scale DNNs on mobile devices.

%algorithm-level innovations and hardware-level accelerations. Algorithm-level innovations should fully take advantage of the optimization knobs at compiler and hardware levels to maximize parallelism. We propose a combination of \emph{pattern pruning} and \emph{connectivity pruning}, exhibiting high degrees in both \emph{flexibility} and \emph{regularity}. The \underline{flexibility} is desirable at theory/algorithm level, matching theoretical results on the desirable kernel shape and principles in human visual systems, resulting in high accuracy. It also enables a key optimization knob, the \emph{re-ordering and code generation} ability of compiler. In this way both instruction-level and thread-level parallelism can be maximized/maintained. The high degree of \underline{regularity} enables another important optimization knob, the \emph{redundant load elimination} technique of compiler, for further hardware performance enhancement. We shall see that our pattern-based DNN acceleration framework, PatDNN, is desirable at all of theory, algorithm, compiler, and hardware levels. 

%triangular memory trade-off (input/output/DNN model)
%and the potential of converting it to a two-side memory trade-off (input/output)

%\textcolor{red}{List of novelty here.}

%\textcolor{red}{The analogy here.}

%\textcolor{red}{A summary of results here.}

%% file: tex/background.tex
\section{Background and Motivation}

\subsection{Layerwise Computation of DNNs}

\begin{figure}%{r}{0.5\textwidth}
    \centering
    \includegraphics[width=0.45 \textwidth]{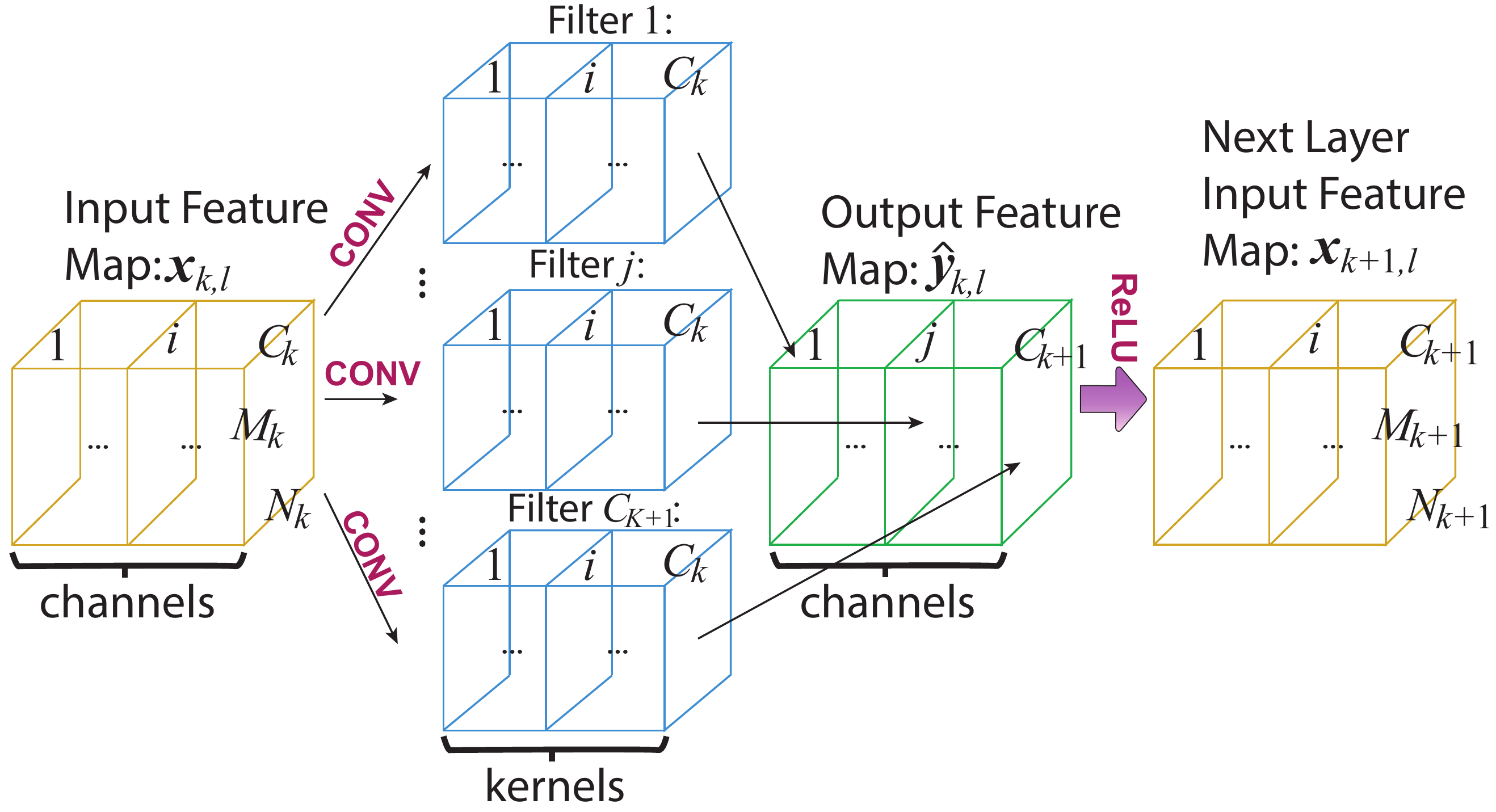}
    %\vspace{-1mm}
    \caption{DNN CONV layer computation.}
    \label{fig:DNNlayer}
    %\vspace{-2mm}
\end{figure}

DNN models can be viewed as cascaded connections of multiple functional layers, such as convolutional (CONV), fully-connected (FC), and pooling (POOL) layers, to extract features for classification or detection~\cite{lee2009convolutional,karpathy2014large,yu2011deep}.
Take the most computation-intensive CONV layer as an example, 
as shown in Figure \ref{fig:DNNlayer}, the input feature map of the $k$-th layer has a size of $M_k \times N_k \times C_k$, where $C_k$ is the number of \emph{channels} of the input feature map.
This layer uses $C_{k+1}$ CONV filters, each with a size of $P_k \times Q_k \times C_k$. 
Note that the number of \emph{kernels} $C_k$ in a CONV filter should match the number of channels $C_k$ in the input feature map to perform convolution.
Each $j$-th CONV filter performs convolution with the input feature map, using a stride of $S_k$, resulting in the $j$-th channel in the output feature map.
Therefore, the number of channels in the output feature map equals to the number of filters $C_{k+1}$, while the size of the output feature map i.e., $M_{k+1}$ and $N_{k+1}$ is determined by $M_k$, $N_k$, $P_k$, $Q_k$, and $S_k$.
The CONV layer is followed by an activation layer, which performs an activation operation, typically ReLU, on the output feature map.
Besides the functional layers in DNNs, {batch normalization} becomes an essential operation to increase the stability of DNN training by overcoming the gradient vanishing issue \cite{ioffe2015batch}. 

\subsection{Mobile Acceleration of DNNs}

In recent years, there have been intensive efforts 
on DNN inference acceleration frameworks targeting mobile devices, include DeepX~\cite{lane2016deepx}, TFLite~\cite{TensorFlow-Lite}, DeepEar~\cite{lane2015deepear}, TVM~\cite{chen2018tvm}, Alibaba Mobile Neural Network (MNN) \cite{Ali-MNN}, DeepCache~\cite{xu2018deepcache},  DeepMon~\cite{huynh2017deepmon}, DeepSense~\cite{yao2017deepsense}, and MCDNN~\cite{han2016mcdnn}. 
Most of these prior works do not fully utilize model compression techniques.
%explore the possible optimization opportunities like computation and memory foot print reductions offered by model compression (e.g., weight pruning). 
%\textcolor{red}{TODO: add DeftNN, CVPR}
Other efforts that explore model sparsity and model compression to accelerate the DNN execution include Liu et al.~\cite{liu2015sparse}, DeftNN~\cite{hill2017deftnn}, SCNN~\cite{parashar2017scnn}, AdaDeep~\cite{liu2018demand}. However, they either do not target mobile platforms, or require new hardware, or trade off compression rate and accuracy,
introducing various drawbacks compared to our work.

Table~\ref{tab:dnn-frameworks} compares the major optimization techniques offered by three state-of-the-art, end-to-end DNN inference frameworks (TFLite~\cite{TensorFlow-Lite}, TVM~\cite{chen2018tvm}, and MNN~\cite{Ali-MNN}).
We do not include other efforts, e.g., DeepCache~\cite{xu2018deepcache} and DeepMon~\cite{huynh2017deepmon}, since they mainly focus on specific DNN applications rather than general DNNs.
In this work, our goal is to find
the most appropriate weight pruning scheme for mobile DNN acceleration and the corresponding full-stack acceleration framework. We utilize 16-bit floating point representation on GPU for both weights and intermediate results which is supported in mobile devices and shown to incur no accuracy loss \cite{TensorFlow-Lite,Ali-MNN,chen2018tvm} for DNNs.
%Our results show that it incurs no accuracy loss.

%both in our proposed framework and baselines. 
%As 8-bit fixed point representation 
%is also supported in current mobile devices,
%the proposed techniques can be also applied
%in this case and may lead to 
%further performance enhancement. 

%is already supported in current mobile devices. This is an orthogonal direction to this work, and can be eventually combined for further performance and energy efficiency enhancement.

\begin{table}
\scriptsize
\caption{DNN acceleration frameworks on mobile devices.}\label{tab:dnn-frameworks}
\vspace{-1.5em}
{\setlength{\tabcolsep}{3.6pt}
\begin{tabular}{llcccc}\\
\toprule
DNNs
 & Optimization Knobs & TFLite   & TVM  & MNN & {\bf Ours}\\
\midrule
& Parameters auto-tuning & N & Y & N & {\bf Y}\\
& CPU/GPU support & Y & Y & Y & {\bf Y}\\
Dense
& Half-floating support & Y & Y & Y & {\bf Y} \\
& Computation graph optimization & Y\tmark[$!$] & Y\tmark[*] & Y\tmark[$!$] & {\bf Y}\tmark[**] \\
& Tensor optimization & Y\tmark[$!$] & Y\tmark[$\dag$] & Y\tmark[$!$] & {\bf Y}\tmark[$\dag\dag$]\\
\midrule
%& Quantization support & Y & Y & Y & {\bf Y} \\

& Sparse DNN model support & N & N & N & {\bf Y}\\
& Pattern-based pruning & N & N & N & {\bf Y}\\
Sparse 
& Connectivity pruning & N & N & N & {\bf Y}\\
& Filter kernel reordering & N & N & N & {\bf Y}\\
& Opt. sparse kernel code generation & N & N & N & {\bf Y}\\
%& Ternary weights & N & N & N & {\bf Y}\\
& Auto-tuning for sparse models & N & N & N & {\bf Y}\\
\bottomrule
\multicolumn{5}{l}{{\tiny * Operator fusion, constant folding, static memory plan, and data layout transform}} \\
\multicolumn{5}{l}{{\tiny ** Besides above in *, operation replacement}} \\
\multicolumn{5}{l}{{\tiny $\dag$ Scheduling, nested parallelism, tensorization, explicit memory latency hiding}} \\
\multicolumn{5}{l}{{\tiny $\dag\dag$ Besides above in $\dag$, dense kernel reordering, SIMD operation optimization}} \\
\multicolumn{5}{l}{{\tiny $!$ Similar optimizations as TVM, but less advanced}}
\end{tabular}
}
% {
% \multicolumn{5}{l}{
%   \begin{minipage}{7.5cm}
%     \tiny * Operator fusion, constant folding, static memory plan, and data layout transform\\
%     \tiny ** Besides above in *, operation replacement\\
%     \tiny $\dag$ Scheduling, nested parallelism, tensorization, explicit memory latency hiding\\
%     \tiny $\dag\dag$ Besides above in $\dag$, dense kernel reordering, SIMD operation optimization\\
%     \tiny $!$ Similar optimizations as TVM, but less advanced\\
%   \end{minipage}%
% }

% % \tnote[*]{Operator fusion, constant folding, static memory plan, and data layout transform}\\
% % \tnote[**]{Besides above in *, operation replacement}\\
% % \tnote[$\dag$]{Scheduling, nested parallelism, tensorization, explicit memory latency hiding}\\
% % \tnote[$\dag\dag$]{Besides above in $\dag$, dense kernel reordering, SIMD operation optimization}\\ 
% %intermediate tensor memory reuse (Q: included in static memory planning ??) 
% % \tnote[$!$]{Similar optimizations as TVM, but less advanced}
% }
\end{table}

\subsection{DNN Model Compression and Challenges}

DNN model compression has been proposed for simultaneously reducing the storage/computation and accelerating inference with minor classification accuracy (or prediction quality) loss. Model compression is performed during DNN training. Two important categories of DNN model compression techniques are weight pruning \cite{han2015learning,guo2016dynamic,dai2017nest,mao2017exploring,wen2016learning,he2017channel} and weight quantization \cite{leng2017extremely,park2017weighted,zhou2017incremental,lin2016fixed,wu2016quantized,rastegari2016xnor,hubara2016binarized,courbariaux2015binaryconnect}.

\begin{figure}%{r}{0.35\textwidth}
    \centering
    \includegraphics[width=0.45 \textwidth]{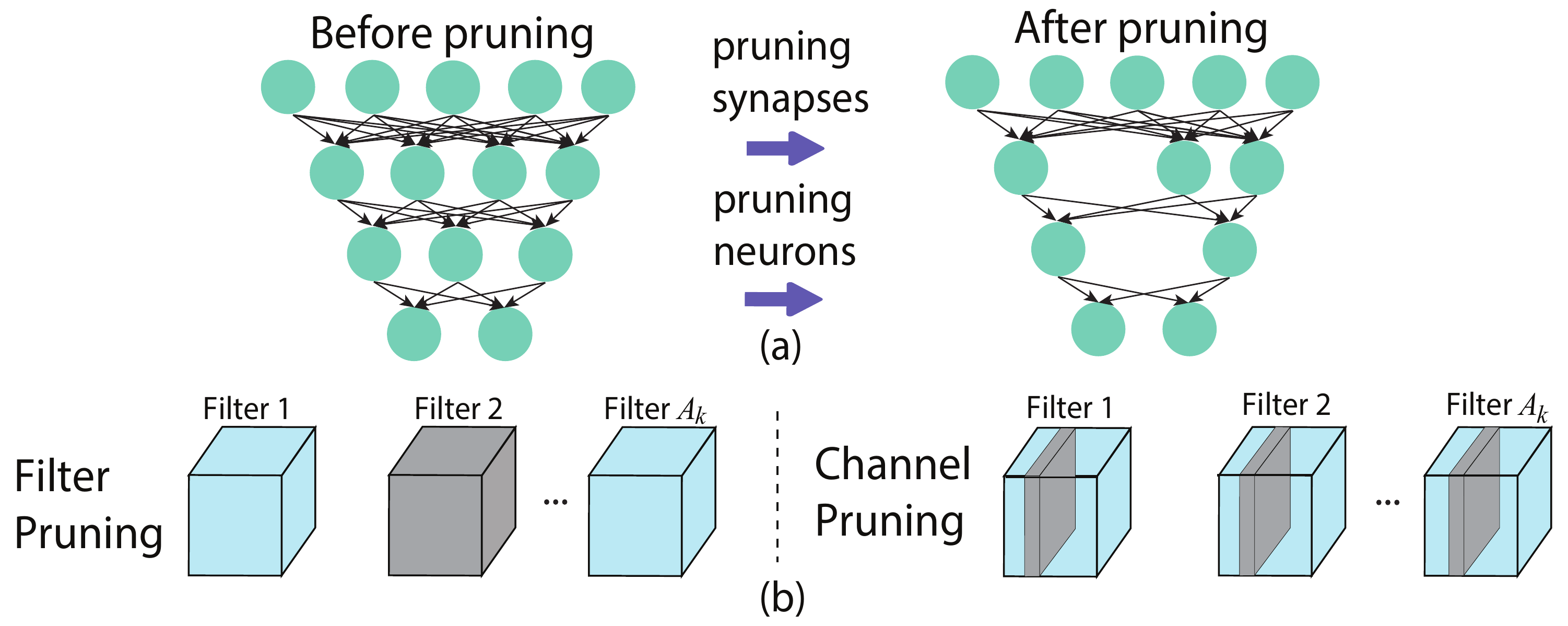}
    % \vspace{-10mm}
    \caption{(a) Non-structured weight pruning and {(b) two types of structured weight pruning.}}
    \label{fig:structuredpruning}
    %\vspace{-3mm}
\end{figure}

{\em Weight pruning} reduces the redundancy in the number of weights.
As shown in Figure~\ref{fig:structuredpruning}, 
two main approaches of weight pruning are
(1) the general and non-structured pruning; and 
(2) structured pruning, which produces
irregular and regular compressed DNN models. 
%the two main approaches \textit{\textbf{non-structured (irregular) weight pruning scheme}} and the \textit{\textbf{structured (regular) weight pruning scheme}} as shown in {Figure \ref{fig:structuredpruning}}.

{\bf Non-Structured Pruning:}
In this method, arbitrary weight can be pruned. 
It can result in a high pruning rate, i.e., 
reduction in the number of weights, 
which can reduce the actual computation. 
For compiler and code optimization, 
non-structured pruning 
incurs several challenges 
due to the irregularity in computation and memory access.
%, thereby resulting in degradation in
%execution performance.
First, the irregular and sparse kernel 
weights require
{\em heavy control-flow instructions}, which 
degrade instruction-level parallelism.
%to handle irregular, sparse kernels that degrade instruction-level parallelism; 
Second, it introduces 
{\em thread divergence and load imbalance} due to the fact that kernels in different filters have divergent workloads and they are usually processed by multiple threads --- a key concern for efficient thread-level parallelism.
Third, it usually incurs {\em low memory performance} due to poor data locality and cache performance.
More importantly, it prohibits advanced memory optimizations such as eliminating redundant loads that widely exist in convolution operations.  
Similarly, for hardware acceleration,
since the pruned models are stored in
some sparse matrix format with indices, 
they often 
lead to performance degradation in GPU and CPU implementations \cite{han2015learning,guo2016dynamic,dai2017nest}. 

% Although non-structured weight pruning can reduce the actual computation, it 

%\subsection{Limitation of Current Structured Pruning}

{\bf Structured Pruning: }
This method can produce regular,
but smaller weight matrices.
{Figure \ref{fig:structuredpruning} (b)} illustrates the representative structured pruning schemes: \emph{filter pruning} and \emph{channel pruning} \cite{wen2016learning}. 
Filter and channel pruning can be considered as equivalent in that pruning a filter in the $k$-th layer is equivalent to pruning the corresponding channel in the $(k+1)$-th layer. Filter/channel pruning is compatible with Winograd algorithm \cite{winograd1980arithmetic,lavin2016fast} 
that has been used to accelerate 
computation of the original DNNs. 
Due to the regular structure, 
the GPU/CPU implementations typically lead to more significant acceleration \cite{mao2017exploring,wen2016learning,he2017channel}.
However, the structured pruning suffers from notable accuracy loss~\cite{wen2016learning,he2017channel}.

\subsection{ADMM-based DNN Model Compression Framework}

Recent work ADMM-NN~\cite{ren2019ADMMNN,ye2019progressive} 
leverages Alternating Direction Methods of Multipliers (ADMM) method for joint DNN weight pruning and quantization. 
ADMM is a powerful tool for optimization, by decomposing an original problem into two subproblems that can be solved separately and efficiently. For example, considering optimization problem $\min_{\bf{x}} f({\bf{x}})+g({\bf{x}}).$ In ADMM, this problem is decomposed into two subproblems on $\bf{x}$ and $\bf{z}$ (auxiliary variable), which will be solved iteratively until convergence. The first subproblem derives $\bf{x}$ given $\bf{z}$: $\min_{\bf{x}} f({\bf{x}})+q_1(\bf{x}|\bf{z})$. The second subproblem derives $\bf{z}$ given $\bf{x}$: $\min_{\bf{z}} g({\bf{z}})+q_2(\bf{z}|\bf{x})$. Both $q_1$ and $q_2$ are quadratic functions.

As a unique property, ADMM can effectively deal with a subset of combinatorial constraints and yield optimal (or at least high quality) solutions~\cite{hong2016convergence,liu2018zeroth}. Luckily, the necessary constraints in the DNN weight pruning and quantization belong to this subset of combinatorial constraints, 
making ADMM applicable to DNN model compression.

Due to the unprecedented results on accuracy 
and pruning rate, ADMM-NN~\cite{ren2019ADMMNN} 
is considered as the state-of-art results for
non-structured weight pruning 
and one of state-of-art methods for weight quantization. 
For non-structured pruning, ADMM-NN achieves 167$\times$, 24$\times$, and 7$\times$ weight reductions on LeNet-5, AlexNet, and ResNet-50 models, respectively, without accuracy loss. However, the framework only focuses on non-structured weight pruning, in which the pruning rate does
not directly translate to performance improvements.

ADMM-NN can be extended to perform structured 
pruning, i.e., filter/channel pruning, and 
our results show that it 
leads to 1.0\% Top-5 accuracy degradation with 3.8$\times$ weight reduction on VGG-16 CONV layers using ImageNet dataset.
Although better than prior work (1.7\% in {\cite{he2017channel}} and 1.4\% in AMC {\cite{he2018amc}}), this accuracy loss is not negligible for many applications.

\subsection{Motivation}

Based on the discussion of prior work
on weight pruning, we rethink the design space 
and observe that non-structured and structured 
represent two extremes in the design space.
In non-structured pruning, any weight can be 
pruned, we consider it as a fine-grained method;
in structured pruning, the weights of
whole filter or channel are pruned together,
we consider it as a coarse-grained method. 
Correspondingly, the two methods have different 
implications on hardware acceleration and 
software optimization: non-structured pruning
is not hardware or software optimization 
friendly, so the higher pruning ratio cannot 
fully translate to performance gain, while 
structured pruning incurs higher accuracy loss. 

The motivation of our study is to seek
an approach that can offer the best of both methods. 
To achieve that, 
we naturally introduce a new dimension,
{\em fine-grained 
pruning patterns inside the coarse-grained
structures}, revealing a previously {\em unknown}
point in design space. 
With the higher accuracy enabled by 
fine-grained pruning pattern, the key question is 
how to re-gain similar hardware efficiency 
as coarse-gained structured pruning. 
We take a unique approach and leverage 
compiler optimizations to 
close the performance 
gap between full structured pruning and 
pattern-based ``semi-structured'' pruning.
%The effectiveness of our approach is 

%between hardware efficiency
%of full structured pruning and the 
%pattern-based ``semi-structured'' pruning.

%% file: tex/overview.tex
\section{Overview of PatDNN}

\subsection{Pattern-based Pruning}

In pattern-based pruning, the key consideration
is how to design and select the patterns. 
To achieve high accuracy and execution efficiency,
we need to design the patterns considering the 
implication for {\em theory and algorithm},
{\em compiler optimization}, and 
{\em hardware execution}.
Good patterns should have two key properties:
flexibility and regularity.

The {\em Flexibility} is not only desirable at theory and algorithm level but also enables efficient
compiler code generation. 
Specifically, it allows compilers to 
maximize or maintain
both instruction-level and thread-level parallelism.
The {\em regularity} not only results in 
highly efficient hardware execution but also 
enables efficient compiler optimizations such as 
{\em redundant load elimination} to 
further improve performance. 
Compared to irregular structures, 
recent works also show from theory and algorithm
level that high accuracy or function approximation
capability can be achieved at the same time 
with certain regularity.
Given these two key properties, 
we propose two pattern-based pruning techniques:
kernel pattern pruning and connectivity pruning.

%The high degree of \underline{regularity} enables another important optimization knob, the \emph{redundant load elimination} technique of compiler, for further enhancing hardware performance. 

%We are in urgent need of a highly accurate, hardware-friendly approach to achieve real-time execution of large-scale DNNs. The \textbf{key motivation} is to use the \underline{compiler} as the key for seamlessly bridging the gap between algorithm-level innovations and hardware-level deployments on mobile devices.

%Specifically, we propose a combination of \emph{pattern pruning} and \emph{connectivity pruning}, exhibiting high degrees in both \emph{flexibility} and \emph{regularity}.

%The \underline{flexibility} is desirable at theory and algorithm level, and also enables a key optimization knob, the \emph{code generation} ability of compiler. Both instruction-level and thread-level parallelism can be maximized or maintained. 
%The high degree of \underline{regularity} enables another important optimization knob, the \emph{redundant load elimination} technique of compiler, for further enhancing hardware performance. We shall see that our pattern-based, end-to-end DNN acceleration framework (PatDNN) becomes theory-algorithm-compiler-hardware all-level desirable.

\begin{figure}%{r}{0.35\textwidth}
    \centering
    \includegraphics[width=0.45 \textwidth]{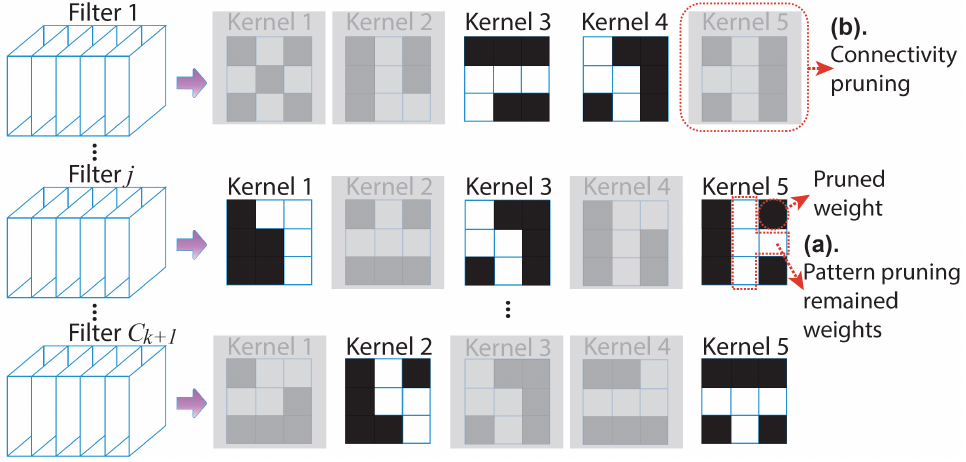}
    %\vspace{-1mm}
    \caption{Illustration of (a) kernel pattern pruning on CONV kernels, and (b) connectivity pruning by removing kernels.}
    \label{fig:pattern_connectivity}
    %\vspace{-3mm}
\end{figure}

\textbf{Kernel Pattern Pruning} is illustrated in {Figure \ref{fig:pattern_connectivity}}. For each kernel (in a CONV filter), a fixed number of weights are pruned, and the remaining weights (white cells) form specific ``kernel patterns''. We define the example in {Figure \ref{fig:pattern_connectivity}} as 4-entry pattern pruning, since every kernel reserves 4 non-zero weights out of the original $3\times 3$ kernel (the most commonly used kernel). 
%Here, we use CONV layer as an example, but 
The same approach is also applicable to other kernel sizes and the FC layer. For each kernel, it possesses \emph{flexibility} in choosing among a number of pre-defined patterns.

At {\em theory and algorithm} level, it is shown in {\cite{li2017pruning,lebedev2016fast}} that the desirable kernel shape has certain patterns 
to match the connection structure in human 
visual systems,
instead of a square shape.
The selection of appropriate pattern for each kernel
can be naturally done by extending 
ADMM-based framework.
In Section~\ref{sec:pattern_acc_result}, we achieve 
accuracy enhancement in all representative 
DNNs in our testing.
At {\em compiler} level, 
the pre-defined pattern allows compiler to 
{\em re-order and generate codes} at filter and kernel level so that kernels with the same pattern
can be grouped for consecutive executions to
maximize instruction-level parallelism.
At {\em hardware} level, the 4-entry patterns are extremely friendly to the SIMD architecture in embedded processors based on either
GPUs or CPUs.
Note that our approach is general and can be 
applied to any pre-defined patterns, not just
the 4-entry considered in the paper.

%When equipped with enhanced ADMM-based framework for selecting appropriate pattern for each kernel, we achieve accuracy enhancement in all representative DNNs in our testing, with details shown in \textcolor{red}{Section xx}. 
%\underline{At compiler level}, the \emph{re-ordering and code generation} technique shall be utilized at kernel level to group the kernels with the same pattern for consecutive executions(\textcolor{red}{Do we need mention filter-level reordering too?}), thereby maximizing instruction-level parallelism.
%\underline{At hardware level}, 4-entry patterns are extremely friendly to the SIMD architecture in embedded processors, both for GPUs and CPUs. 

\begin{figure}%{r}{0.4\textwidth}
    \centering
    \includegraphics[width=0.45 \textwidth]{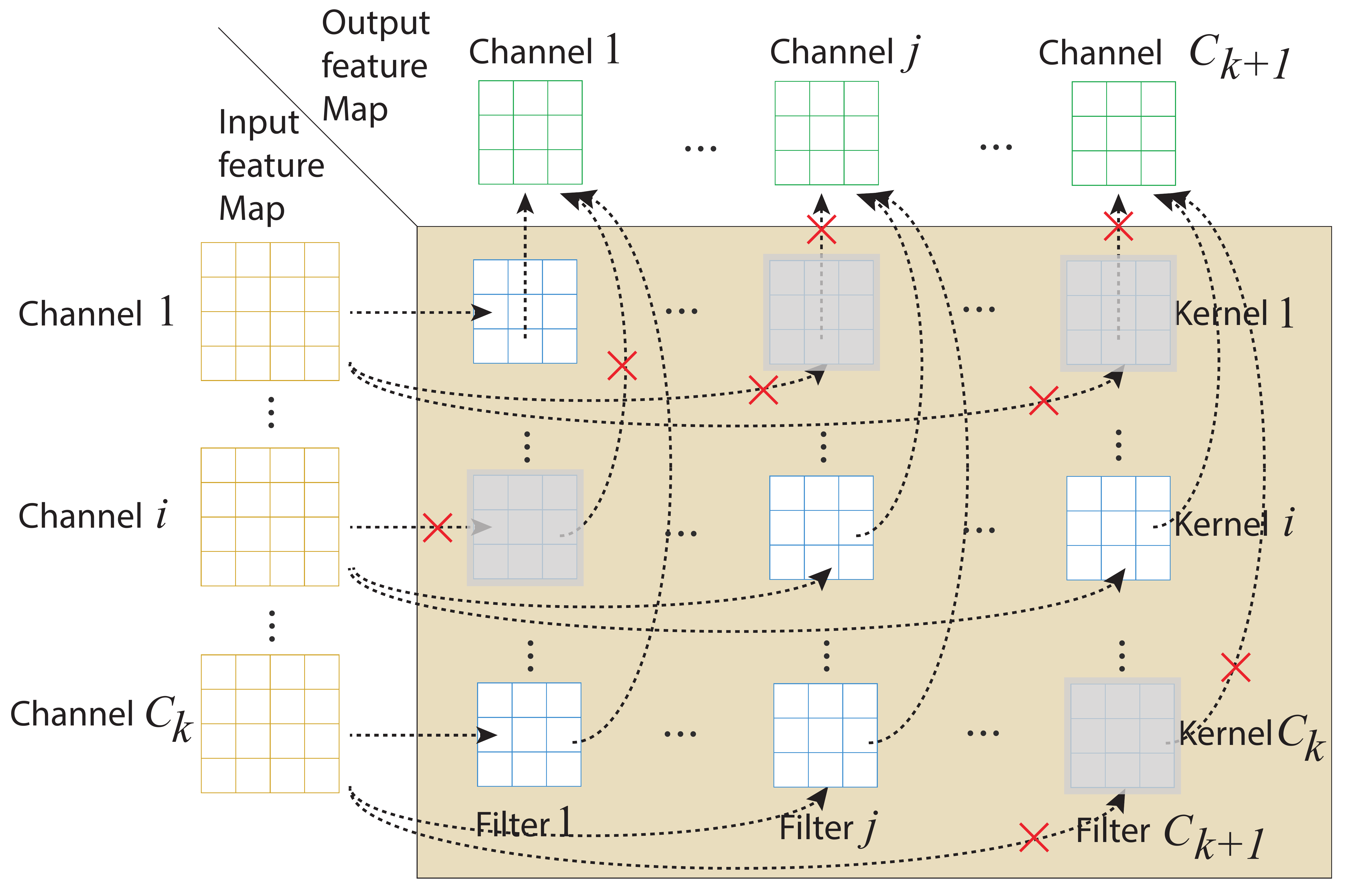}
    %\vspace{-5mm}
    \caption{Illustration of connectivity pruning.}
    \label{fig:connectivity}
    %\vspace{-3mm}
\end{figure}

\textbf{Connectivity Pruning} is illustrated in {Figure \ref{fig:connectivity}}.
The key insight is to {\em cut the connections}
between certain input and output channels, which is equivalent to removal of corresponding kernels. 
In CONV layers, the correlation between input channel $i$ and output channel $j$ is represented by the $i$-th kernel of filter $j$.
%, as shown in {Figure \ref{fig:connectivity}}. 
This method is proposed for overcoming the limited weight pruning rate by kernel pattern pruning.

%by shading a number of pruned/removed kernels for each filter. 
%In CONV layers, the correlation between input channel $i$ and output channel $j$ is represented by the $i$-th kernel of filter $j$, as shown in {Figure \ref{fig:connectivity}}. 

%Connectivity pruning is termed for cutting the connections between certain input and output channels, which is equivalent to removal of corresponding kernels. It is proposed for overcoming the limited weight pruning rate by pattern pruning.

At {\em theory and algorithm} levels, the connectivity pruning matches the desirability of locality in layerwise computations inspired by human visual systems \cite{yamins2016using,yamins2014performance}. It is more flexible than the prior filter/channel pruning schemes that remove whole filters/channels, thereby achieving higher accuracy. At {\em compiler and hardware} level, removed kernels and associated computations can be grouped by compiler using the \emph{re-ordering} capability without affecting the other computations, thereby maintaining parallelism degree.

% \begin{table}%{R}{0.50\textwidth}
% \caption{Qualitative comparison of different pruning schemes on accuracy and speedup under the same pruning rate.}\label{tab:prunecompare}
% %\vspace{-2mm}
% \centering
% \includegraphics[width =  0.9\linewidth]{fig/Table_PruneCompare.pdf}
% \end{table}

\begin{table}%{R}{0.50\textwidth}
\vspace{-5mm}
\caption{Qualitative comparison of different pruning schemes on accuracy and speedup under the same pruning rate.}\label{tab:prunecompare}
\centering
\includegraphics[width =  1\linewidth]{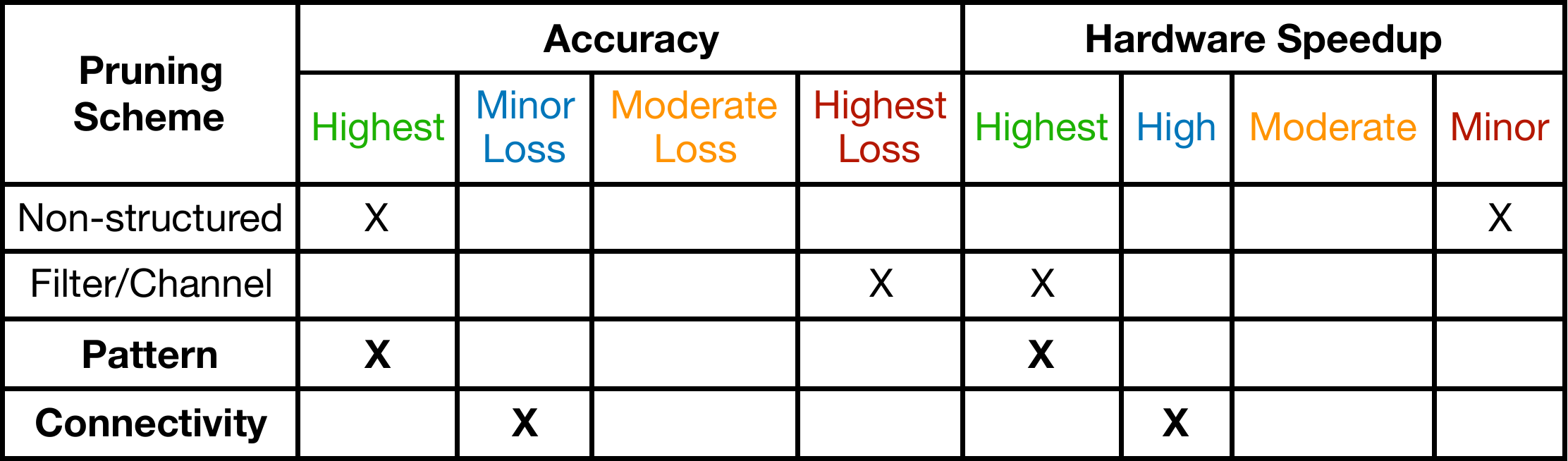}
%\vspace{5mm}
\end{table}

%is not simply a tradeoff of non-structured pruning and filter/channel pruning, where the former is accurate but not hardware friendly, and the latter is opposite.
%The \textbf{key merit} is that PatDNN can exploit the advantages while hiding the weaknesses, becoming both highly accurate and hardware friendly. 
%This is enabled by compiler. 
%A better analogy is the \emph{cache hierarchy} which is both fast (similar to first-level cache) and has high capacity (similar to main memory or disk), enabled by appropriate cache management policy. 

\subsection{Overview of PatDNN Acceleration Framework}

%\todo{an overview figure}

\begin{figure*}%{r}{0.5\textwidth}
    \centering
    \includegraphics[width=1 \textwidth]{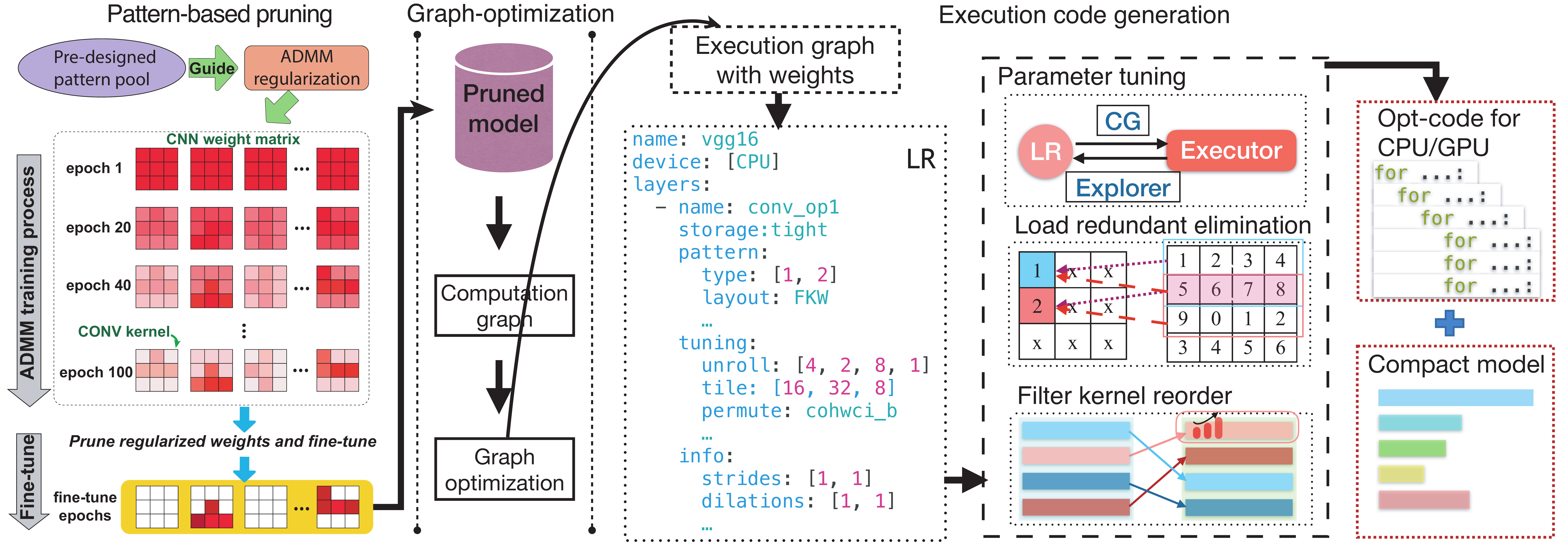}
    \caption{Overview of PatDNN acceleration framework.}
    \label{fig:system-overview}
\end{figure*}

%\textcolor{red}{TODO: support both CPU/GPU}

Based on the above discussions, we propose
{\em \projectname}, a novel end-to-end mobile DNN acceleration framework that can generate highly accurate
DNN models using pattern-based pruning methods
and guarantee execution efficiency with 
compiler optimizations. 
Compared to recent prior works \cite{ren2019ADMMNN,he2018amc,he2017channel,wen2016learning},
\projectname uniquely enables 
{\em cross-layer vertical integration}, 
making it desirable across theory/algorithm,
compiler and hardware. 
Allowing compilers to treat 
pruned kernels as special patterns, 
our approach not only achieves high pruning rate
with high accuracy, 
but also effectively converts into performance 
improvements due to hardware friendly properties. 
%The pruned kernels can be treated 
%as a special pattern by compilers. 
%In this way we can achieve high pruning (acceleration) rate while maintaining hardware friendliness.

As shown in {Table \ref{tab:prunecompare}}, \projectname can achieve the benefits of both non-structured and structured pruning. 
The key enabler to achieving this goal
is to leverage compiler to maintain the 
efficiency of structured pruning based on 
kernel pattern and connectivity pruning. 
Our approach is an excellent example of 
hardware and software co-design, which can be 
compared to an intuitive analogy:
the multi-level cache memory hierarchy provides
sufficient hardware supports to hide memory
access latency and explore locality, but 
compiler and software optimizations are still needed
to fully realize effective 
cache management policy. 

Figure~\ref{fig:system-overview} 
shows the overview of \projectname which consists of
two stages:
(1) {\em pattern-based training stage} (Section~\ref{sec:training}), which
performs kernel pattern and connectivity pruning
with an extended ADMM solution framework. 
(2) {\em execution code generation stage} (Section~\ref{sec:inference}), which
performs multiple effective optimizations based
on the patterns. 
Similar to TVM \cite{chen2018tvm}, \projectname converts DNN models into computational graphs and applies multiple graph-based optimizations. 
%including all ones in TVM and XX. 
Based on these optimizations, we focus on
layerwise design and optimization including 
a high-level and fine-grained DNN layerwise representation (LR), filter kernel reorder, load redundancy eliminations, and automatic parameter tuning. All of these designs and optimizations are general, 
and applicable to both mobile CPUs and GPUs.
%on optimized data placement on mobile GPU,  and loop tiling and permutation. 
The second stage generates optimized execution codes as well as DNN models with weights stored in a novel compact format. 

%% file: tex/pattern-prune.tex
\section{PatDNN Training w/ Pattern-based Pruning}\label{sec:training}

\begin{figure}[t]%{r}{0.5\textwidth}
    \centering
    \includegraphics[width=0.45 \textwidth]{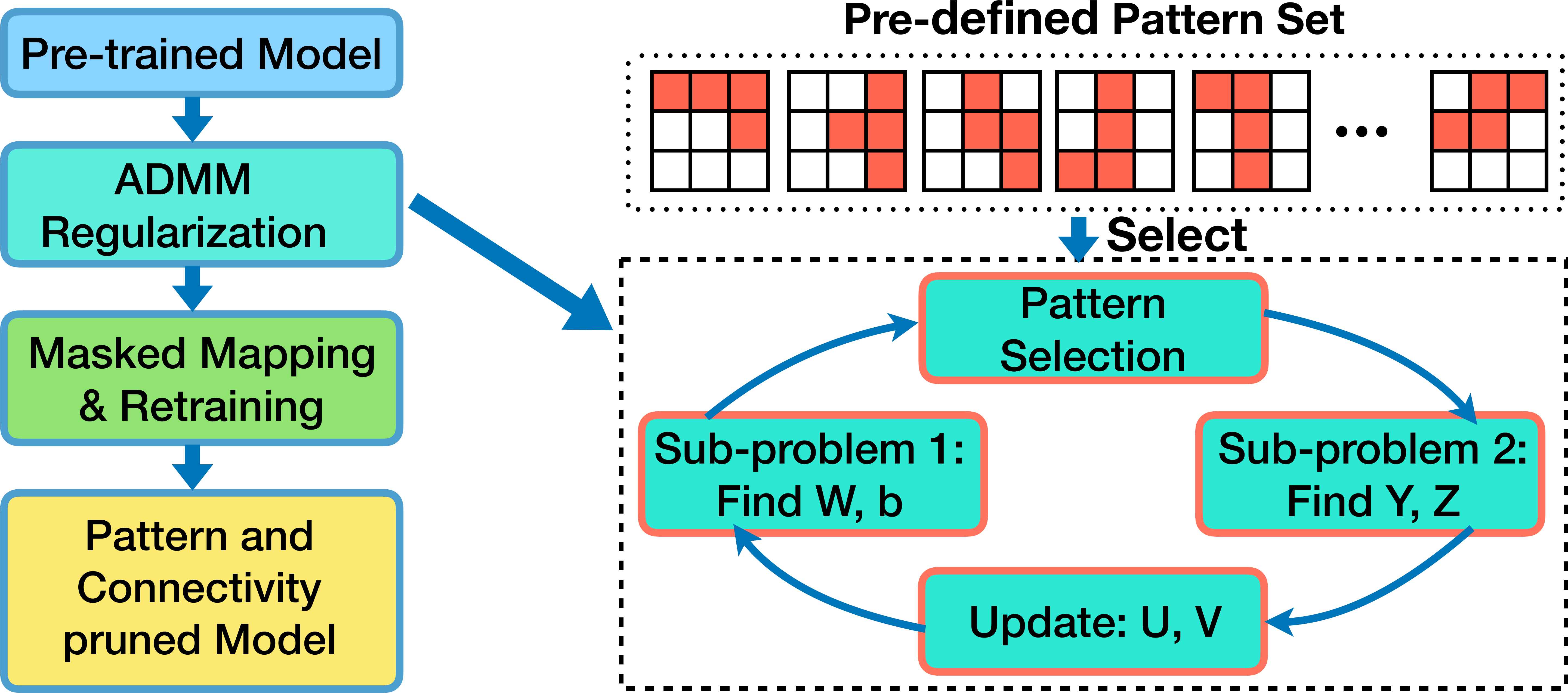}
    %\vspace{-1mm}
    \caption{The algorithm-level overview of PatDNN training.}
    \label{fig:algorithm-overview}
    %\vspace{-2mm}
\end{figure}

This section describes the methods to generate
compressed DNN models for \projectname. 
The procedure is composed of two steps:
(1) we design a set of desired patterns to be 
%such that a desirable pattern can be 
selected for each kernel;
(2) assign a pattern for each kernel (kernel pattern pruning) or prune the whole kernel (connectivity pruning), and train the pattern-based weights for maintaining accuracy. 
The overall flow is shown in Figure~\ref{fig:algorithm-overview}.
Essentially, it reflects the algorithm aspects of 
\projectname.
Our method can be applied to either
a pre-trained DNN or train a model 
from scratch. 

%This section focuses on the algorithm level of PatDNN. We are given a pre-trained DNN, or we can train from scratch. 

%First, we need to design a set of patterns, such that a desirable pattern can be selected for each kernel from the pre-defined pattern set. Next, we need to (i) assign a pattern for each kernel (kernel pattern pruning) or prune the whole kernel (connectivity pruning), and (ii) train remaining weights for maintaining accuracy. The overall flow is shown in Figure~\ref{fig:algorithm-overview}.

\subsection{Designing the Pattern Set}

We need to determine the number of patterns, and design each specific candidate pattern in the pattern set. The number of patterns is an important hyperparameter that should be carefully considered. 
If it is too large, it is more challenging to 
generate efficient codes, thereby affecting 
performance;
if it is too small, the lack of flexibility may lead to 
accuracy degradation. 
%n over-large number will result in difficulty in code generation for compilers and long generated codes, thereby execution overhead, and (ii) an over-small number will result in lower degree of flexibility and perhaps accuracy degradation. 
Through empirical study, we validate that 6-8 patterns in the set achieves as a desirable tradeoff for the most common $3\times 3$ kernel---ensuring low compiler overhead while maintaining high accuracy.
%compiler overhead is low while accuracy can be maintained.

When the number of patterns is determined and 4-entry patterns are utilized, the compiler optimization
and hardware efficiency are oblivious to the 
specific pattern shapes. 
%the specific pattern shapes 
%will not affect the compiler and hardware performance. 
However, the specific patterns to use need to be 
carefully optimized to maintain high accuracy
after kernel pattern pruning. 
%they will affect
%the accuracy of models after 
%kernel pattern pruning and need to be
%will be affected and shall be 
%optimized through designing specific patterns. 
The key insights of pattern design are: 
(1) both theory and empirical studies~\cite{yamins2014performance,yamins2016using} show that the central weight in a $3\times 3$ kernel is critical and shall not be pruned; and 
(2) it is desirable that the distortion is small for each kernel before and after kernel pattern pruning. Hence, we propose the following heuristic. 
First, for the pre-trained DNN, we scan all the kernels, and for each kernel, we find the four weights with largest magnitudes (including the central weight). These four weights form a 4-entry pattern, called the \emph{natural pattern} of the kernel. According to the definition of natural patterns, there are a total of $\binom{8}{3}=56$ number of possible patterns.
Suppose we aim at $k$ different patterns in the candidate set. We count and select the Top-$k$ most commonly appeared natural patterns across all kernels in the DNN, thereby forming the pattern candidate set (to select from in the subsequent step).

Our study on pattern number and pattern style selection is consistent with the pattern pruning theory work that is proposed in ~\cite{ma2019pconv}. Different from pattern theory derivation in~\cite{ma2019pconv}, our approach focuses on system-level design and compiler optimization of the pattern-based acceleration framework. 
%In this way we have designed the pattern set.

\subsection{Kernel Pattern and Connectivity Pruning Algorithm}

\textbf{Problem Formulation:} Consider an $N$-layer DNN, and we focus on the most computationally intensive CONV layers. The weights and biases of layer $k$ are respectively denoted by ${\bf{W}}_{k}$ and ${\bf{b}}_{k}$, and the loss function of DNN is denoted by $f \big( \{{\bf{W}}_{k}\}_{k=1}^N, \{{\bf{b}}_{k} \}_{k=1}^N \big)$, refer to \cite{zhang2018systematic}
for more details. In our 
discussion, $\{{\bf{W}}_{k}\}_{k=1}^N$ and $\{{\bf{b}}_{k} \}_{k=1}^N$ respectively characterize the collection of weights and biases from layer $1$ to layer $N$. Then the pattern and connectivity pruning is formulated as an optimization problem:
\begin{equation}
\label{opt0}\small
\begin{aligned}
& \underset{ \{{\bf{W}}_{k}\},\{{\bf{b}}_{k} \}}{\text{minimize}}
& & f \big( \{{\bf{W}}_{k}\}_{k=1}^N, \{{\bf{b}}_{k} \}_{k=1}^N \big),
\\ & \text{subject to}
& & {\bf{W}}_{k}\in {\mathcal{S}}_{k},\ {\bf{W}}_{k}\in {\mathcal{S}}'_{k},\ \; k = 1, \ldots, N.
\end{aligned}
\end{equation}
The collection of weights in the $k$-th CONV layer forms a four-dimensional tensor, i.e., ${\bf{W}}_{k} \in R^{P_k \times Q_k \times C_k \times C_{k+1}}$, where $P_k, Q_k, C_k$, and $C_{k+1}$ are respectively the height of kernel, the width of kernel, the number of kernels, and the number of filters, in layer $k$. Suppose $\bf{X}$ denotes the weight tensor in a specific layer, then $({\bf{X}})_{:,:,a,b}$ denotes a specific kernel. 

In \emph{kernel pattern pruning}, the constraint in the $k$-th CONV layer is ${\bf{W}}_{k}\in {\mathcal{S}}_{k}
:=
\{{\bf{X}}\mid$ {each kernel in} ${\bf{X}}$ {needs to satisfy one specific pattern shape in the pattern set (and non-zero weight values can be arbitrary)}$\}$.
In \emph{connectivity pruning}, the constraint in the $k$-th CONV layer is ${\bf{W}}_{k}\in {\mathcal{S}}'_{k}
:=
\{{\bf{X}}\mid$ {the number of nonzero kernels in} ${\bf{X}}$ {is less than or equal to} $\alpha_k \}$ ($\alpha_k$ is a predetermined hyperparameter with more discussions later).
Both constraints need to be simultaneously satisfied.

\textbf{Extended ADMM-based Solution Framework:} The constraint ${\bf{W}}_{k}\in {\mathcal{S}}_{k}$ in problem (\ref{opt0}) is different from the clustering-like constraints in ADMM-NN \cite{ren2019ADMMNN}, in that it is flexible to select a pattern for each kernel from the pattern set. As long as a pattern is assigned for each kernel, constraints in problem (\ref{opt0}) become clustering-like and ADMM compatible.
Similar to ADMM-NN \cite{ren2019ADMMNN}, the ADMM-based solution is an iterative process, starting from a pre-trained DNN model. We assign an appropriate pattern for each kernel based on the $L_2$-norm metric in each iteration, to achieve higher flexibility.

By incorporating auxiliary variables ${\bf{Z}}_{k}$'s and ${\bf{Y}}_{k}$'s, and dual variables ${\bf{U}}_{k}$'s and ${\bf{V}}_{k}$'s, we decompose (\ref{opt0}) into three subproblems, and iteratively solve until convergence.
In iteration $l$, after assigning patterns we solve the first subproblem\vspace{-0.1in}
\begin{align}\label{eqn:5}\small \nonumber
 &\underset{ \{{\bf{W}}_{k}\},\{{\bf{b}}_{k} \}}{\text{minimize}}
\ f \big( \{{\bf{W}}_{k} \}_{k=1}^N, \{{\bf{b}}_{k} \}_{k=1}^N \big)+\sum_{k=1}^{N} \frac{\rho_{k}}{2}  \| {\bf{W}}_{k}-{\bf{Z}}_{k}^{l}+{\bf{U}}_{k}^{l} \|_{F}^{2}\vspace{-0.1in}\\
&+  \sum_{k=1}^{N} \frac{\rho_{k}}{2}  \| {\bf{W}}_{k}-{\bf{Y}}_{k}^{l}+{\bf{V}}_{k}^{l} \|_{F}^{2}.
\vspace{-0.5in}\end{align}
The first term is the loss function of the DNN, while the other quadratic terms are convex. 
As a result, this subproblem can be solved by stochastic gradient descent (e.g., the ADAM algorithm \cite{kingma2014adam}) similar to training the original DNN.

The solution $\{{\bf{W}}_{k}\}$ of subproblem 1 is denoted by $\{{\bf{W}}^{l+1}_{k}\}$. Then we aim to derive $\{{\bf{Z}}^{l+1}_{k}\}$ and $\{{\bf{Y}}^{l+1}_{k}\}$ in subproblems 2 and 3. These subproblems have the same form as those in ADMM-NN \cite{ren2019ADMMNN}. Thanks to the characteristics in combinatorial constraints, the optimal, analytical solution of the two subproblems are Euclidean projections, and are polynomial time solvable. For example, for connectivity pruning, the projection is: keeping $\alpha_k$ kernels with largest $L_2$ norms and setting the rest of kernels to zero. For kernel pattern pruning it is similar.
Finally, we update dual variables ${\bf{U}}_{k}$ and ${\bf{V}}_{k}$ according to the ADMM rule \cite{boyd2011distributed} and thereby complete the $l$-th iteration in the ADMM-based solution.

The hyperparameter determination process is relatively straightforward for joint pattern and connectivity pruning. There is no additional hyperparameters for kernel pattern pruning when the pattern set has been developed. For connectivity pruning we need to determine the pruning rate $\alpha_k$ for each layer. In this paper, we adopt a heuristic method of uniform pruning rate for all layers except for the first layer (which is smaller, yet more sensitive to pruning). 

\subsection{Accuracy Validation and Analysis}\label{sec:pattern_acc_result}

We validate the accuracy of ADMM-based joint kernel pattern and connectivity pruning, based on ImageNet ILSVRC-2012 and CIFAR-10 datasets, using VGG-16 \cite{simonyan2014very}, ResNet-50 \cite{he2016deep}, and MobileNet-V2 \cite{sandler2018mobilenetv2} DNN models.
Our implementations are based on PyTorch, and the baseline accuracy results are in many cases higher than prior work, which reflects the recent 
progress in DNN training. With a pre-trained DNN model, we limit the number of epochs in kernel pattern and connectivity pruning to 120, similar to the original DNN training in PyTorch and much lower than iterative pruning \cite{han2015learning}. 

\begin{table}[h]
\vspace{2mm}
\caption{Top-5 accuracy comparison on kernel pattern pruning.}
\label{tab:pattern_acc_result}
\centering
\includegraphics[width = 1\linewidth]{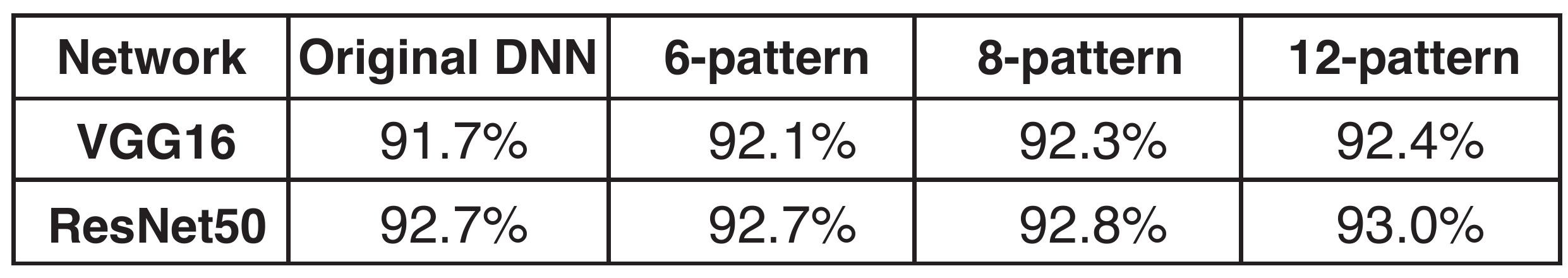}
%\vspace{0.5mm}
\end{table}

Table~\ref{tab:pattern_acc_result} illustrates the Top-5 accuracy comparison on kernel pattern pruning only, applied on the CONV layers of VGG-16 and ResNet-50 using ImageNet dataset. The baseline is the original DNN without patterns, and we demonstrate the accuracy results with 6, 8, and 12 patterns (all 4-entry patterns) in the pattern set. 
Our first observation is that \emph{the accuracy will improve when the number of candidate patterns is sufficient} --- typically 4 - 8 patterns are sufficient. This is attributed to the compatibility of kernel pattern pruning with human visual system and the ability to eliminate overfitting (compared with square kernel shape). This observation has been also
validated for other types of DNNs and data sets (e.g., CIFAR-10).

\begin{table}[h]
\vspace{2mm}
\caption{Top-5 accuracy and CONV weight reduction on joint kernel pattern pruning (8 patterns in the set) and connectivity pruning.}
\label{tab:8pattern_compare}
\centering
\includegraphics[width = 1\linewidth]{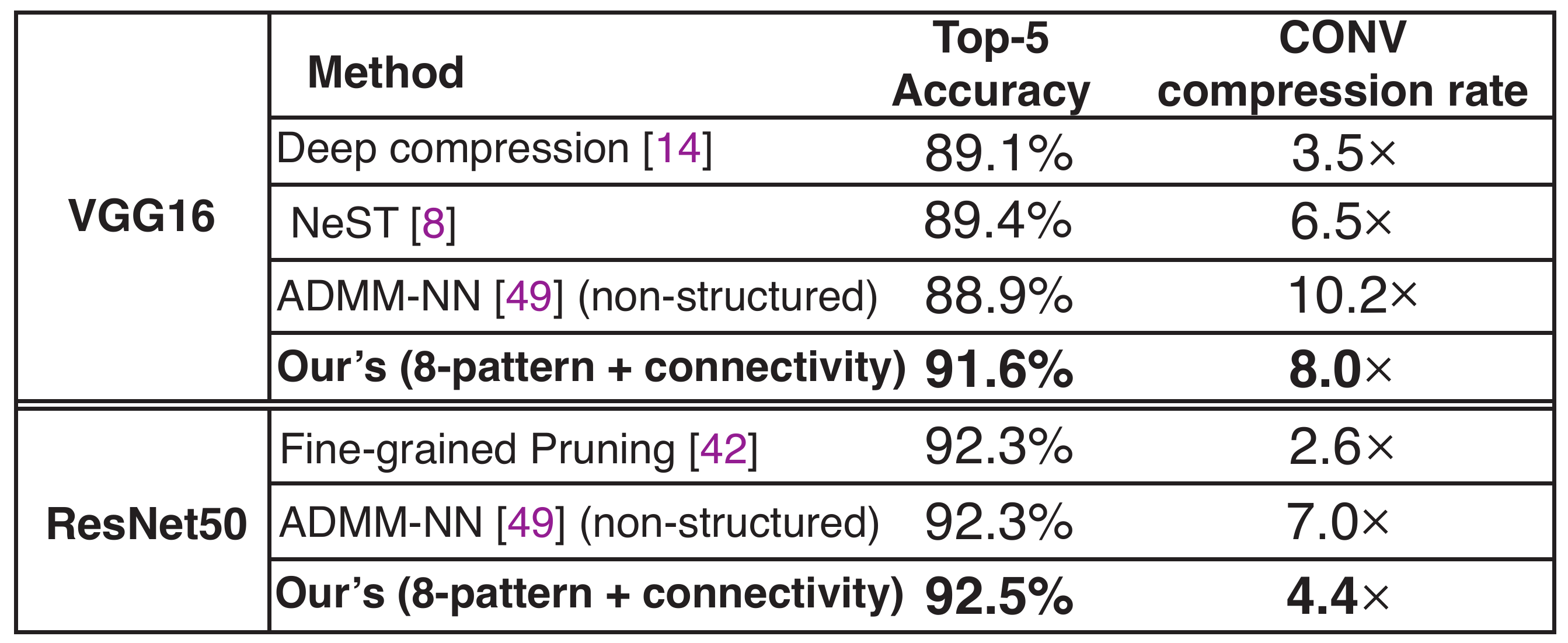}
\vspace{1mm}
\end{table}

Table~\ref{tab:8pattern_compare} illustrates the Top-5 accuracy comparison on joint kernel pattern pruning (8 patterns in the set) and connectivity pruning, on VGG-16 and ResNet-50 using ImageNet dataset. For VGG-16, all kernels are $3\times 3$. After applying 4-entry patterns on all kernels and 3.6$\times$ uniform connectivity pruning, we achieve around 8$\times$ weight reduction on CONV layers of VGG-16. For ResNet-50, a portion of kernels are $1\times 1$ besides the majority of $3\times 3$ kernels. We apply kernel pattern pruning on all $3\times 3$ ones, and apply uniform 3.6$\times$ connectivity pruning on all kernels. We achieve 4.4$\times$ weight reduction on CONV layers. One can observe from the table that (1) \emph{no Top-5 accuracy drop with this setup}; (2) \emph{under the same accuracy, the weight reduction rate is close to ADMM-based (and outperforms prior heuristic based) non-structured pruning on CONV layers.}

For the CIFAR-10 dataset, we observe consistent accuracy improvements with 8 patterns on 3$\times$3 kernels and 3.6$\times$ connectivity pruning, with results shown in Section \ref{sec:evaluation}.

%% file: tex/design.tex
\section{PatDNN Inference Code Optimization}\label{sec:inference}

For DNN models with kernel pattern and connectivity pruning, \projectname ensures hardware execution
efficiency of DNN inference
with optimized compiler and code generation. 
%relies on a compiler and code generation based stage to optimize DNN models and corresponding execution code. 
As aforementioned, compiler optimizations
play the key role in ``recovering'' the performance
loss due to the fine-grained pattern-based
pruning compared to fully structured pruning. 
%
%the pattern-based DNN execution still suffers from the performance issues of non-structured weight pruning. 
%
This stage includes two-levels of optimizations: (1) optimizations on computational graphs that explore the potential opportunities among multiple DNN layers;
and (2) optimizations within each layer. \projectname adopts an enhanced TVM~\cite{chen2018tvm}-like approach together with other innovations from the latest efforts in this direction (e.g., Tensor Comprehensions~\cite{vasilache2018tensor}) to implement the former (with major optimizations summarized in Table~\ref{tab:dnn-frameworks}). 
Due to space limit, we do not elaborate each
as they are not the main research contribution
and not specific to DNN execution optimization 
leveraging pattern-based pruning.

%This is not the main research contribution of PatDNN or not specific to pattern-based pruned DNN executions, thus not further elaborated due to space constraints. 

This section focuses on \projectname's layerwise optimizations based on kernel pattern and connectivity pruning 
%that are specifically aimed to improve the execution performance of sparse DNNs with pattern-based pruning. 
that are specifically designed to address the challenges in DNN acceleration with non-structured weight pruning, i.e., {\em heavy control-flow instructions, thread divergence and load imbalance, and poor memory performance}. These optimizations are general, and applicable to both mobile CPUs and GPUs. Our framework can generate both optimized CPU (vectorized C++) code and GPU (OpenCL) code. 
%\blue{
Figure~\ref{fig:compiler-flow} illustrates PatDNN's compiler-based optimization and code generation flow with a CONV layer example. 
%} 
%Particularly, this section will demonstrate that our pattern-based pruning design brings multiple good opportunities of resolving these challenges.
%, i.e., heavy control-flow instructions, thread divergence and load imbalance, and poor memory performance.

\subsection{Compiler-based PatDNN Inference Framework}\label{sec:ir}

\begin{figure*}[t]%{r}{0.5\textwidth}
    \centering
    \includegraphics[width=0.9 \textwidth]{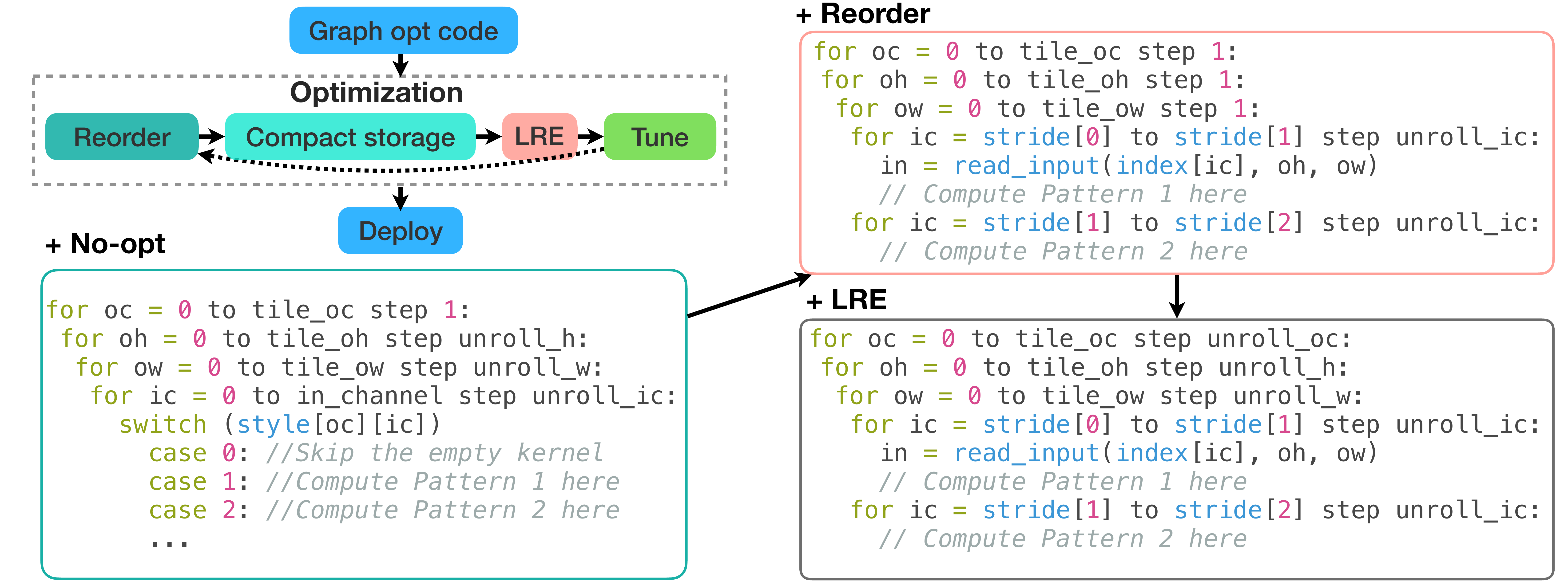}
    \vspace{2mm}
    \caption{{\bf PatDNN's compiler-based optimization and code generation flow:} compiler takes both model codes with graph-based optimizations and a layerwise representation (as an example in Figure~\ref{fig:ir-example}) to generate low-level C/C++ and OpenCL codes (as {\tt No-opt}). This low-level code is further optimized with filter kernel reorder and our FKW compact model storage ({\tt +Reorder}), the register-level load redundancy elimination ({\tt +LRE}), and other optimizations like auto-tuning. Finally, the code is deployed on mobile devices.}
    \label{fig:compiler-flow}
    %\vspace{-2mm}
\end{figure*}

\begin{figure}%{r}{0.5\textwidth}
    \centering
    \includegraphics[width=0.45 \textwidth]{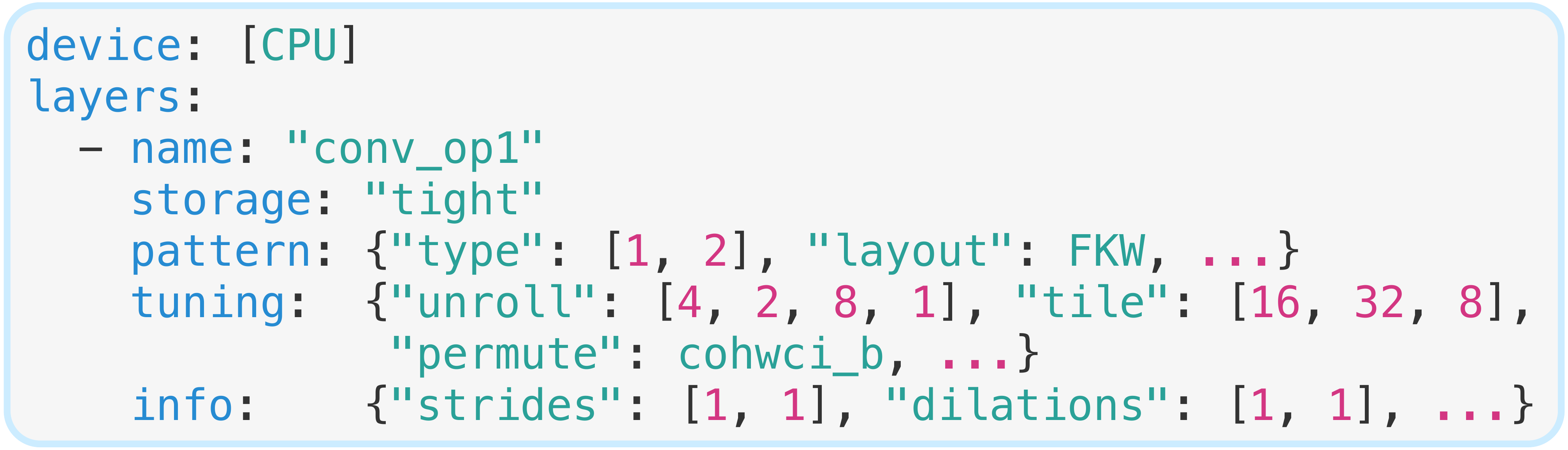}
    \vspace{2mm}
    \caption{An LR example for a CONV layer.}
    \label{fig:ir-example}
    %\vspace{-2mm}
\end{figure}
% \begin{figure}%{r}{0.5\textwidth}
%     \centering
%     \includegraphics[width=0.5 \textwidth]{fig/CoCiHW.pdf}
%     %\vspace{-1mm}
%     \caption{CoCiHW computation pattern - Wei(Sketch).}
%     \label{fig:filterchannel}
%     %\vspace{-2mm}
% \end{figure}
% \begin{figure}%{r}{0.5\textwidth}
%     \centering
%     \includegraphics[width=0.5 \textwidth]{fig/CoHWCi.pdf}
%     %\vspace{-1mm}
%     \caption{CoHWCi computation pattern - Wei(Sketch).}
%     \label{fig:filterchannel}
%     %\vspace{-2mm}
% \end{figure}
% Based on an IR for each layer
% General compiler approach: model analysis based on IR, IR optimization, and code generation.
% Similar to TVM, inherit its advantages, and design new optimization for pattern based pruning specifically. Our framework also supports DNN models from varied sources like TVM.
% original code, loop permutation, loop blocking

%\blue{Revise: clear explanation and more detailed example}

\noindent{\bf Layerwise Representation:} 
%\blue{
The key feature of \projectname
is its {\em sparsity- and pruning-aware} design. To support it, \projectname proposes a high-level fine-grained Layerwise Representation (LR) to capture
the sparsity information.
%} %Compared with TVM's IR, 
This LR includes intensive DNN layer 
specific information to enable aggressive layerwise optimizations.
In particular, it includes detailed {\em kernel pattern and connectivity-related information} (e.g., the pattern types presented in this layer, the pattern order in each filter, the connection between kernels and input/output channels, etc.); and {\em tuning-decided parameters} (e.g., the input and output tile sizes, unrolling factors, the loop permutation of this layer, etc.).   

\projectname extracts the pattern/connectivity information from DNN models with computational graph optimizations, and determines the tuning-related
parameters by the auto-tuning. 
%As shown in Figure~\ref{fig:compiler-flow}, 
This LR is used for \projectname's following optimizations:
(1) filter kernel reordering, which operates
on kernel pattern and connectivity-related 
information, i.e., specifically the compressed weight storage structure;  
% (2) load redundancy elimination, which
% requires each kernel's pattern; and 
% (3) increasing data reuse, which is 
% determined by the connection between kernels,  input/output channels, and the exact input/output tile sizes and unroll factors. 
and (2) load redundancy elimination, which
requires each kernel's pattern, the connectivity between kernels and input/output channels, and the exact input/output tile size and unroll factor. %\blue{
After these optimizations, high-level LR 
can generate compressed model and associated optimized model execution code by using the pattern-related information and other basic layer information extracted from DNN models, (e.g., the kernel size, computation stride, computation dilation, etc).
Figure~\ref{fig:compiler-flow} shows the optimization flow and two sample code skeletons ({\tt +Reorder} and {\tt +LRE}) for these two optimizations, respectively.%}

%by using the information related to 
%kernel patterns, connectivity and tuning
%together with other basic layer information extracted from DNN models, e.g., the kernel size, computation stride, computation dilation, etc. %this high-level IR can generate compressed model and associated optimized model execution code.  

Figure~\ref{fig:ir-example} shows a simplified LR example for a CONV layer (with 2-D kernels). This LR will generate execution code for CPU ({\tt device}). Two types of kernel patterns ({\tt [1, 2]}) present in this layer ({\tt patterns}) and the filter kernels' pattern layout is specified by our {\tt FKW} compressed weight storage format (clarified in Section~\ref{sec:format} in detail)\footnote{This LR is used after our filter kernel reorder, so the pattern information is stored in the optimized FKW format. Before reorder, a relatively loose data format is used, which is omitted due to the space limit.}. Its computation loop permutation is {\tt cohwci\_b}, i.e., in the order of output channel, output height, output width, and input channel, with blocking and unrolling. Their blocking sizes are specified in {\tt tile}. Their unrolling factors are specified in {\tt unroll}.  Figure~\ref{fig:compiler-flow} ({\tt +Reorder}) also shows the execution code generated from this LR, in which the outer loops iterating on all tiles are omitted. The inner-most iteration processes kernels in each filter in the order of their pattern types, i.e., all kernels with pattern 1 in each filter will be processed at first, then kernels with pattern 2. This code optimization does not require any loop control-flows. This is guaranteed by our filter kernel reorder that is introduced in Section~\ref{sec:reorder} in details.

\subsection{Filter Kernel Reorder (FKR)}\label{sec:reorder}

\begin{figure}%{r}{0.5\textwidth}
    \centering
    \includegraphics[width=0.48 \textwidth]{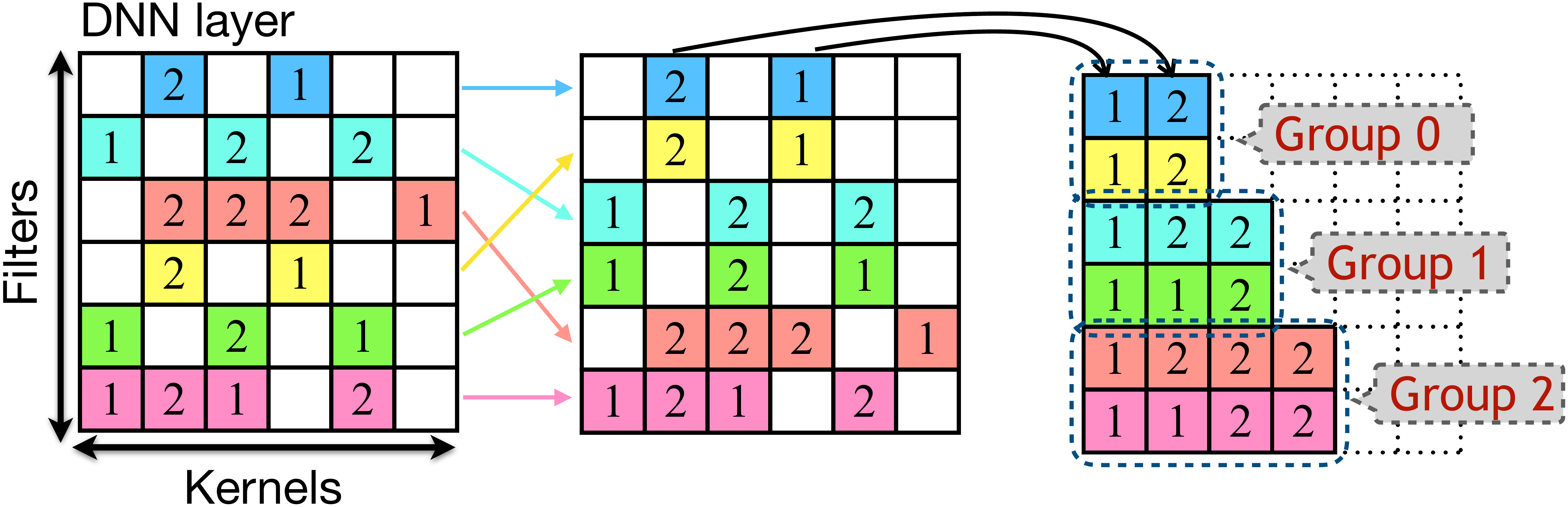}
    %\vspace{2mm}
    \caption{An example of filter kernel reorder.}
    \label{fig:kernel-reorder}
    %\vspace{-2mm}
\end{figure}

%remove branch operations by loop split and group the same pattern together
Kernel pattern and connectivity pruning offer better opportunities to address the performance challenges in non-structured pruning thanks to its better regularity.
Specifically, 
Filter kernel reorder (FKR) is designed to address two key challenges, i.e., heavy control-flow instructions, and thread divergence and load imbalance. Our basic insight is: for a specific DNN layer, the patterns of all kernels are already known after model training, so the inference computation pattern is also known before model deployment. FKR leverages this knowledge to organize the filters with similar kernels together to improve {\em inter-thread} parallelization and order the same kernels in a filter together to improve {\em intra-thread} parallelization.   

Figure~\ref{fig:kernel-reorder} explains FKR with a simplified example. Here, a matrix represents a CONV layer of DNN and each cell is a kernel with pattern type denoted by the number on it. Empty kernels are the ones pruned by {\bf connectivity pruning}. The kernels in the same row belong to the same filter, and are marked with the same color.

Before the reorder, kernels with different patterns are distributed in this DNN layer. When performing the convolution operation directly, the execution code will contain many branches (%\blue{
as the {\tt +No-opt} code in Figure~\ref{fig:compiler-flow})
%} 
that incur significant instruction pipeline stalls and thread divergences, hurting both instruction- and thread-level parallelism. According to our experimental results in Section~\ref{sec:evaluation}, this version results in sub-optimal performance.%, even worse than dense computation.

FKR is composed of two steps: {\em filter reorder} and {\em kernel reorder}. The filter reorder organizes similar filters next to each other and the kernel reorder groups kernels with identical patterns in each filter together. Particularly, the {\em filter similarity} used in filter reorder is decided by two factors: first, the number of non-empty kernels in each filter (i.e., the length of each filter); and second, for filters with the same length, the number of kernels at identical positions with identical pattern IDs when the kernels in each filter are ordered according to these IDs.   

After the reorder, the filters with the same length are grouped together, and in each group, the filters with the highest degree of similarity are ordered next to each other. 
%The code on the right-hand side of Figure~\ref{fig:reorder-code} 
%\blue{
The code {\tt +Reorder} in figure~\ref{fig:compiler-flow} 
is for the execution of this reordered layer.
%} 
This code shows much better instruction-level parallelism because it eliminates all branches. In addition, it also allows the better exploration of thread-level parallelism, because it results in large thread execution similarity and good load balance, particularly, considering the example of mapping the filters in the same group to the same GPU thread block.

%{\textcolor{red}{mention order data?}}

\subsection{Compressed DNN Weight Storage (FKW Format)}\label{sec:format}
\begin{figure}%{r}{0.5\textwidth}
    \centering
    \includegraphics[width=0.48 \textwidth]{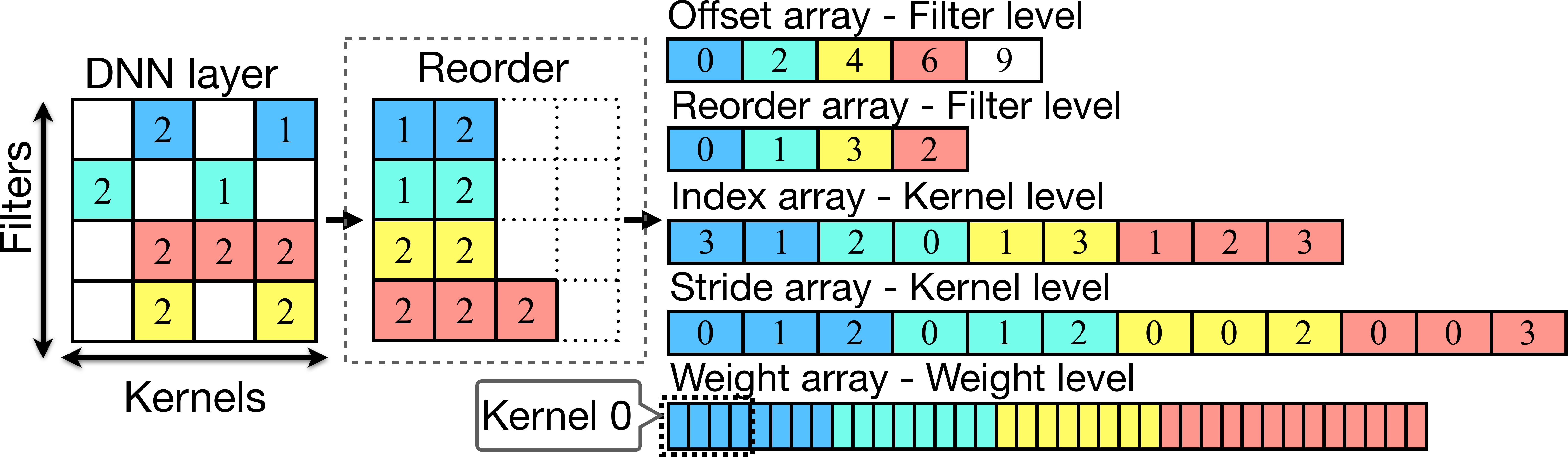}
    %\vspace{-1mm}
    \caption{An example of FKW compressed weight storage.}
    \label{fig:fkw-example}
    %\vspace{-2mm}
\end{figure}

% \begin{figure}%{r}{0.5\textwidth}
%     \centering
%     \includegraphics[width=0.5 \textwidth]{fig/tight_vs_csr.pdf}
%     %\vspace{-1mm}
%     \caption{Tight VS CSR(50X less extra memory needed) - Wei(Sketch).}
%     \label{fig:fkw-csr}
%     %\vspace{-2mm}
% \end{figure}

\begin{figure*}[th]
%\begin{subfigure}{\textwidth}
    \includegraphics[width=0.45\textwidth]{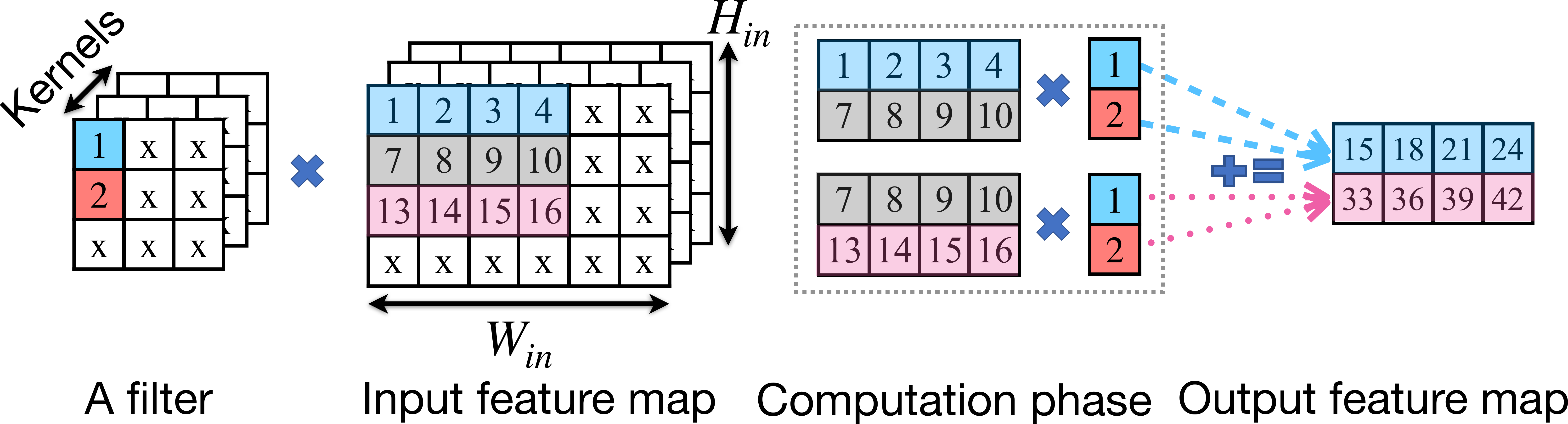}
    %\caption{Kernel Level}
    %\label{figure:re-kernel-level}
%\end{subfigure}
\hfill
%\begin{subfigure}{\textwidth}
    \includegraphics[width=0.45\textwidth]{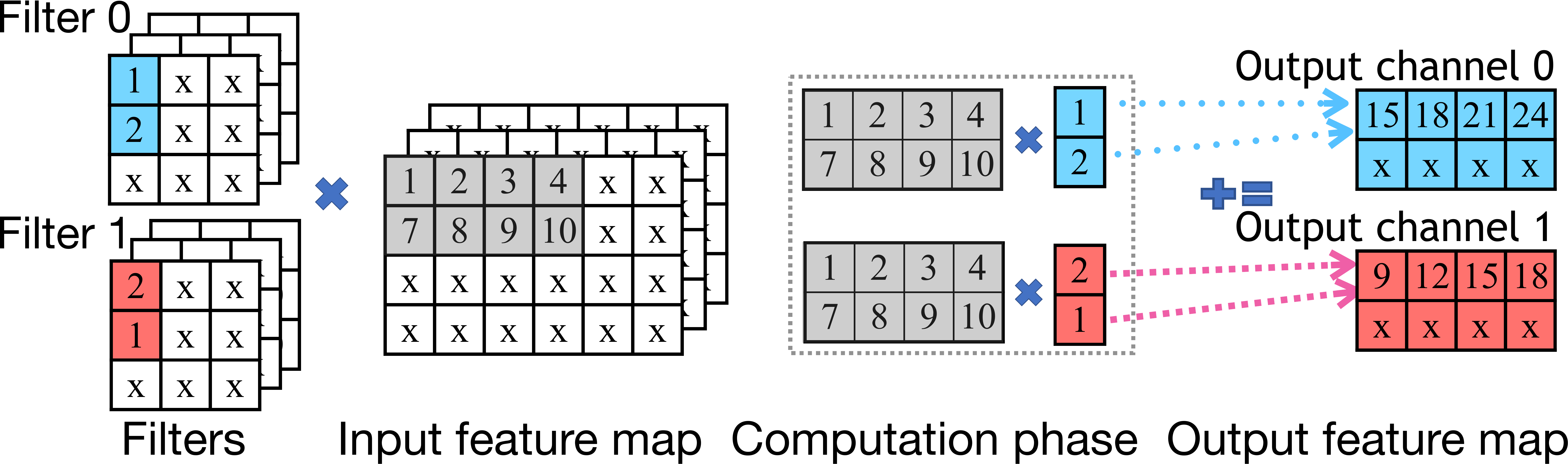}
    %\caption{Filter Level}
    %\label{figure:re-filter-level}
%\end{subfigure}
\caption{Load redundancy elimination (left: kernel-level; right: filter-level).}
\label{figure:load-redundancy-elimination}
    \vspace{-0.5em}
\end{figure*}

After FKR, our LR stores the DNN's weights in a novel compact format (called FKW, standing for Filter-Kernel-Weight format). Compared with existing compact data formats (like CSR), FKW is higher-level and results in much less extra structure overhead (i.e., %{\color {blue} %the size of data structures other than weights ???
the total size of all index arrays that are used for weights data access%}
). 
%\textcolor{red}{I think less overhead is better than better compression rate. Because you do not reduce the number of weights here.} 
In addition, FKW leverages the pattern information, and stores the kernels with the FKR information that will support later branch-less DNN execution. Other compact data format cannot support this.  

Figure~\ref{fig:fkw-example} shows an example. This DNN layer consists of four filters, each with 2, 2, 2, and 3 (after FKR) 
%\textcolor{red}{before or after reordering?} 
non-empty kernels, respectively. The two kernels in the first filter (marked as blue) have pattern 1 and 2, corresponding to the input channel 3 and 1, respectively. FKW uses five arrays to represent this DNN layer: offset array, reorder array, index array, stride array, and weight array. The offset array and reorder array store filter-level information, index array and stride array store kernel-level, and the weight array stores actual weights.

More specifically, the offset array stores the offset of each filter (in terms of the number of non-empty kernels). In Figure~\ref{fig:fkw-example}, the offset of filter 0 is 0, and the offset of filter 1 is 2 because there are two kernels in filter 0, and so on. 
The reorder array shows the reorder information that is used for accumulating the computation output to the correct output channel. In Figure~\ref{fig:fkw-example}, the reorder array tells us that filter 2 and filter 3 have been switched and their computation results should also be switched to the corresponding output channel. %\textcolor{red}{right means correct?} 
The index array represents the corresponding input channel for each non-empty kernel. In Figure~\ref{fig:fkw-example}, kernel 1 in filter 0 %\textcolor{red}{before reorder or after?} 
corresponds to the input channel 3, and kernel 2 corresponds to the input channel 1. So, the first two elements in the index array are 3 and 1, respectively. 
The stride array denotes the number of kernels in each pattern within the same filter. In Figure~\ref{fig:fkw-example}, the filter 0 has the stride array values 0, 1, and 2, denoting that
the filter 0 has 
%0 kernels with pattern 0 ($0 = 0 - 0$), 
%\textcolor{red}{Actually we don't need to point out Kernel 0, because we will skip this automatically, we can say we have pattern 1 start from 0 and end to 1}, 
1 kernel with pattern 1 ($1 = 1 - 0$), and 1 kernel with pattern 2 ($1 = 2 - 1$).
In this example, each kernel has four (non-zero) weights, so each filter has 8, 8, 8, and 12 weights (after FKR), respectively. %\textcolor{red}{This is confusing, before or after reordering?}, respectively.

% \textcolor{red}{TODO: explain the example and compare to csr}
% Figure~\ref{fig:fkw-csr} compares the storage cost of FKW with CSR, theoretically. \textcolor{red}{TODO: big-O analysis}

\subsection{Load Redundancy Elimination (LRE)}\label{sec:redundancy}

% \begin{figure*}[th]
% \begin{subfigure}{0.45\textwidth}
%     \includegraphics[width=0.45\textwidth]{fig/load_redundant_elimination_in_kernel.pdf}
%     \caption{Kernel Level}
%     \label{figure:re-kernel-level}
% \end{subfigure}\hfill
% \begin{subfigure}{0.45\textwidth}
%     \includegraphics[width=0.45\textwidth]{fig/load_redundant_elimination_in_filter.pdf}
%     \caption{Filter Level}
%     \label{figure:re-filter-level}
% \end{subfigure}
% \caption{Load Redundancy Elimination}
% \label{figure:load-redundancy-elimination}
% \end{figure*}

% \begin{figure*}[!ht]
%     \centering
%         \subfloat[Kernel-Level]{\includegraphics[width=0.95\columnwidth]{fig/redundant-kernel.pdf}}
%         \qquad
%         \subfloat[Filter-Level]{\includegraphics[width=0.95\columnwidth]{fig/redundant-filter.pdf}}
%         \caption{Load Redundancy Elimination}
%     \label{fig:subfigname}
% \end{figure*}

% \begin{figure*}[th]
% \begin{subfigure}{0.4\textwidth}
%     \includegraphics[width=0.4\textwidth]{fig/load_redundant_elimination_in_kernel.pdf}
%     \caption{Kernel Level}
%     \label{figure:re-kernel-level}
% \end{subfigure}
% \hfill
% \begin{subfigure}{0.4\textwidth}
%     \includegraphics[width=0.4\textwidth]{fig/load_redundant_elimination_in_filter.pdf}
%     \caption{Filter Level}
%     \label{figure:re-filter-level}
% \end{subfigure}
% \caption{Load Redundancy Elimination (Left: Kernel-Level; Right: Filter-Level)}
% \label{figure:load-redundancy-elimination}
% \end{figure*}

As discussed before, irregular memory access (in the form of array indirection) is also a major cause of inefficient execution of weight pruned DNNs.   
PatDNN uses two techniques to address this issue: 
(1) a conventional input tiling to improve the cache performance; and 
(2) the optimized code generation
with the help of the pre-defined pattern information.
%{\color {blue} %==> The first one is described in ...??
The first one, specifically the determination of the optimal tiling size  will be introduced in Section~\ref{sec:auto-tuning}. 
%} 
This section focuses on the second, specifically, introducing our novel redundant {\em register} load elimination optimization applied in code generation procedure.   

% Our model pruning is able to compress the model size, thus improving the memory and cache performance. 

Our key insight is: in DNN execution, such as a convolution operation, the data access pattern of the input and output is decided by the (none-zero elements) patterns of kernels that are already known after training. Therefore, it is possible to generate the optimized data access code with this information for each pattern of kernels and call them dynamically during the DNN execution. The generated codes consist of all statically determined data access instructions for the kernel-level computation with a careful instruction reorganization to 1) eliminate all indirect memory accesses; and 2) eliminate all redundant {\em register} load operations.
The elimination of all indirect memory accesses is relatively straightforward, because in all data access instructions, the index of input data can be directly calculated from kernel pattern. We next explain two novel register-level load redundancy elimination methods in details.

% Our pattern-based method also allows us to eliminate the redundant {\em register} load operations, therefore optimizing the register performance. Please note such redundancy elimination is impossible for random prune approaches because of the irregular memory access.  

% Two kinds of redundancy: 1. convolution operation redundancy; 2. multiple filters use the same input

% sensitive to tiling size. large tiling results in more redundancy elimination

Figure~\ref{figure:load-redundancy-elimination} illustrates both register-level load redundancy eliminations: the left one is within each kernel, and the right one is among multiple kernels. Within each kernel, the load redundancy is caused by the convolution operation. In the example (shown on the left part of Figure~\ref{figure:load-redundancy-elimination}), the kernel value 1 requires the elements in the first two rows of the input matrix while value 2 requires the second and third rows. The elements in the second row $[7, 8, 9, 10]$ are loaded twice (from cache to register). PatDNN eliminates this load redundancy by explicitly reusing the (SIMD) registers that already hold the required data (like the second row in the above example).

Multiple kernels on the same position of different filters may share the same pattern and input channel.
%---this is very common considering PatDNN has a limited number of patterns while a DNN layer usually contains hundreds of kernels. 
The input data required by these kernels are exactly identical. The right-hand side of Figure~\ref{figure:load-redundancy-elimination} shows a concrete example. If the computation of these filters on identical data is packed together, the possible redundant load of this input can be eliminated. PatDNN explores this optimization when it generates the optimized memory access code. The FKR organizes the kernels (in different filters) with identical patterns together. Together with a filter-level (or output channel) loop unrolling when processing these kernels, the redundant register load is eliminated. %\blue{
Figure~\ref{fig:compiler-flow} ({\tt +LRE}) shows an example of this unrolling code.
%} %\textcolor{red}{Note by Wei, here maybe confusing}. 

% These two load redundancy issues broadly exist in DNN convolution operations, therefore, our optimization yields significant performance benefit that is shown in our evaluation section.

%\blue{
It is worth noting that the above two redundancy elimination opportunities are more straightforward to exploit for dense models where the memory accesses of kernel weights are continuous and the data reuse pattern is periodically repeated. However, it is very challenging (or even not possible) to exploit for pruned sparse models with irregular memory accesses, because it is hard to detect the data reuse pattern (or the data reuse pattern does not even exist). Our pattern-based pruning can preserve the data reuse patterns and help the compiler to detect them, thus re-enabling these two kinds of register-level load redundancy elimination.
%}

%% file: tex/implementation.tex
%\section{Further Implementation and Optimization}

\subsection{Parameter Auto-tuning}\label{sec:auto-tuning}

Many configuration parameters require careful tuning to guarantee the performance of the generated execution code. However, manual tuning is tedious, and hard to yield the optimal code.
Therefore, PatDNN also includes an auto-tuning component for selecting the best execution configuration.

It consists of two parts: first, an {\em explorer model} based on Genetic Algorithm to generate the configuration exploration space; and second, a {\em performance estimation model} created from our historical data to predict the possible best configuration and performance for a given hardware. 
Compared with the simulated annealing in TVM, our explorer model supports better parallelism because it allows the initialization of an arbitrary number of chromosomes to start the search.
%rather than run algorithm in different threads or multiple times. 
For a typical (large-scale) DNN like VGG-16, our exploration can complete in 3-5ms. During the exploration, history data is also collected for training the performance estimator (based on Multilayer Perceptron and least square regression loss). The advantage of this approach is that when deploying PatDNN on a new platform, it can give a quick prediction of the optimal configuration parameters as well as the possible execution time.
In addition, these tuning parameters are crucial to the performance of our PatDNN execution, thus need
to be carefully tuned by our auto-tuning module
including: data placement configurations on GPU, tiling sizes, loop permutations, and loop unrolling factors. 
%\subsection{Efficient Input/Output Tiling}
% \begin{itemize}[leftmargin=*,noitemsep]
% \item {\bf Data Placement on GPU:}
% %shared memory on GPU; and other memory
% Mobile GPU contains various memory devices, texture memory (L1 cache) that is used only for image data, L2 cache that is shared by every kind of data, and unified main memory that is shared with other processors. Placing proper data on right memory is critical to the execution performance.   
% % \item {\bf SIMD and Register Optimization:}
% % vectorization and register opt
% \item {\bf Tiling and Loop Permutation:} Different platforms (CPUs, GPUs, and even different SoCs) may have different memory/cache configurations and capacities, so the best tile size, and loop order may be varied. They also require carefully tuning.  
% \end{itemize}
% \subsection{Model Quantization}
% \todo{we may remove this part and change the whole section to automatic parameter selection}

% \subsection{Other Optimizations}
% \subsection{IR example}
%\input{tex/drafts.tex}

%% file: tex/evaluation.tex
\begin{table}[t]
\caption{DNNs characteristics (under kernel pattern and connectivity pruning): Accu: ImageNet top-5, CIFAR top-1; the negative values in Accuracy Loss actually mean accuracy improvement.}
\label{tab:datasets}
\centering
\includegraphics[width=1\linewidth]{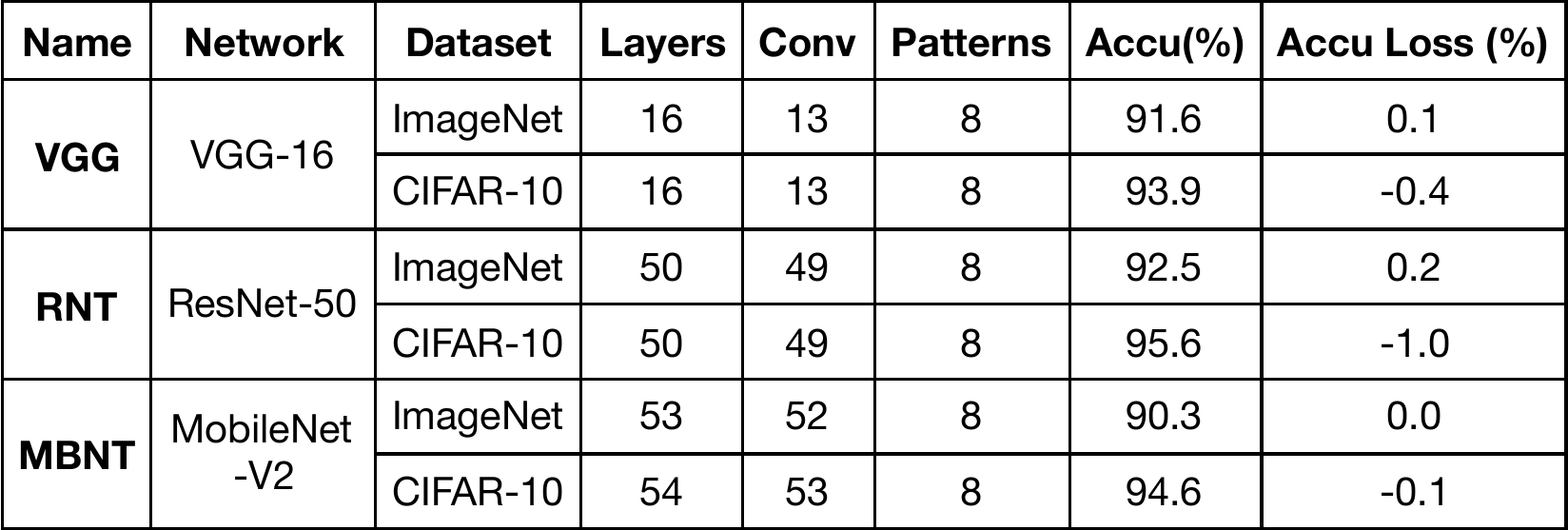}
%\vspace{0.1cm}
\end{table}

\begin{table}[t]
\caption{VGG unique CONV layers' filter shapes and given names.}
\label{tab:eva_vgg_shortname}
\centering
\includegraphics[width = 1\linewidth]{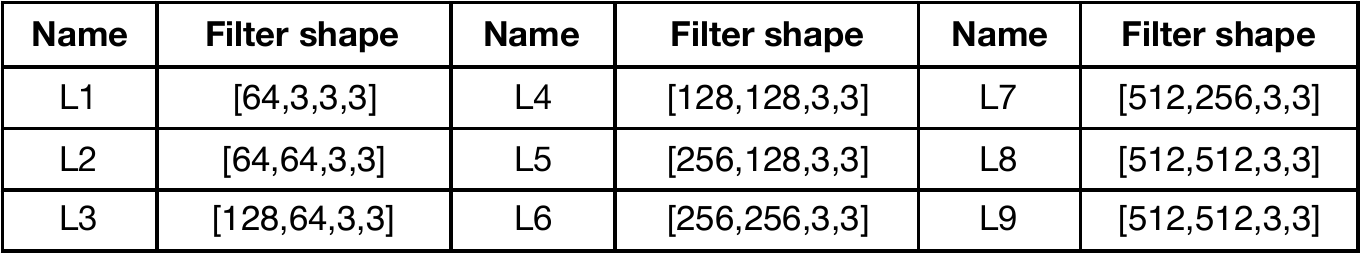}
\vspace{0.1cm}
\end{table}

%\vspace{0.5cm}
\section{Evaluation}\label{sec:evaluation}

This section evaluates the execution performance of PatDNN by comparing it with three state-of-the-art DNN inference acceleration frameworks, 
% TensorFlow Lite (TFLITE)~\cite{TensorFlow-Lite}, TVM (TVM)~\cite{chen2018tvm}, and Alibaba Mobile Neural Network (MNN)~\cite{Ali-MNN}.
TFLite~\cite{TensorFlow-Lite}, TVM~\cite{chen2018tvm}, and MNN~\cite{Ali-MNN}. 
All major optimizations of these frameworks (and our PatDNN) are summarized in Table~\ref{tab:dnn-frameworks}.   

\begin{figure*}[t]
    \centering
        \subfloat[ImageNet-CPU]{
            \includegraphics[width=0.235\textwidth]{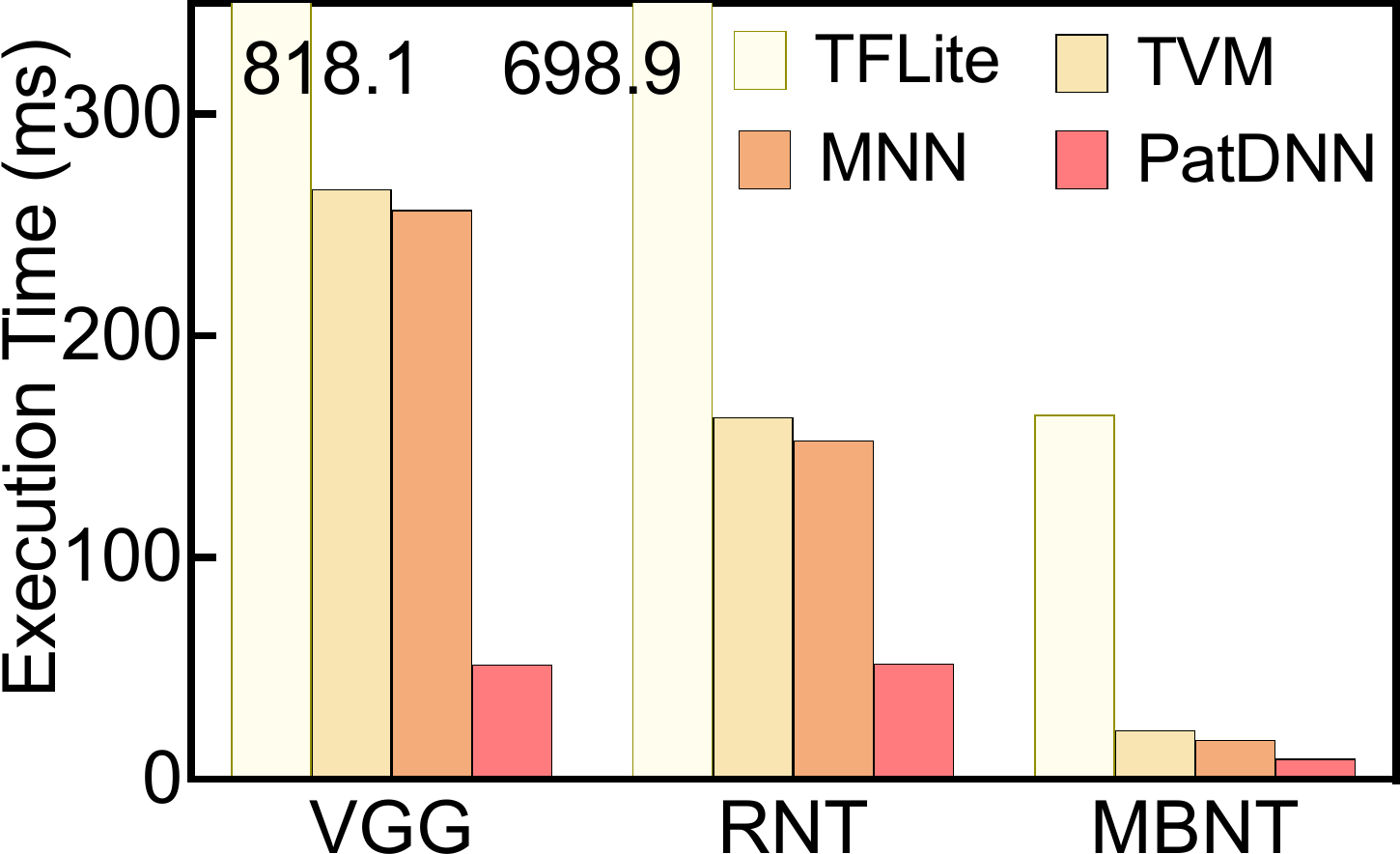}
        }
        \subfloat[CIFAR-10-CPU]{
            \includegraphics[width=0.25\textwidth]{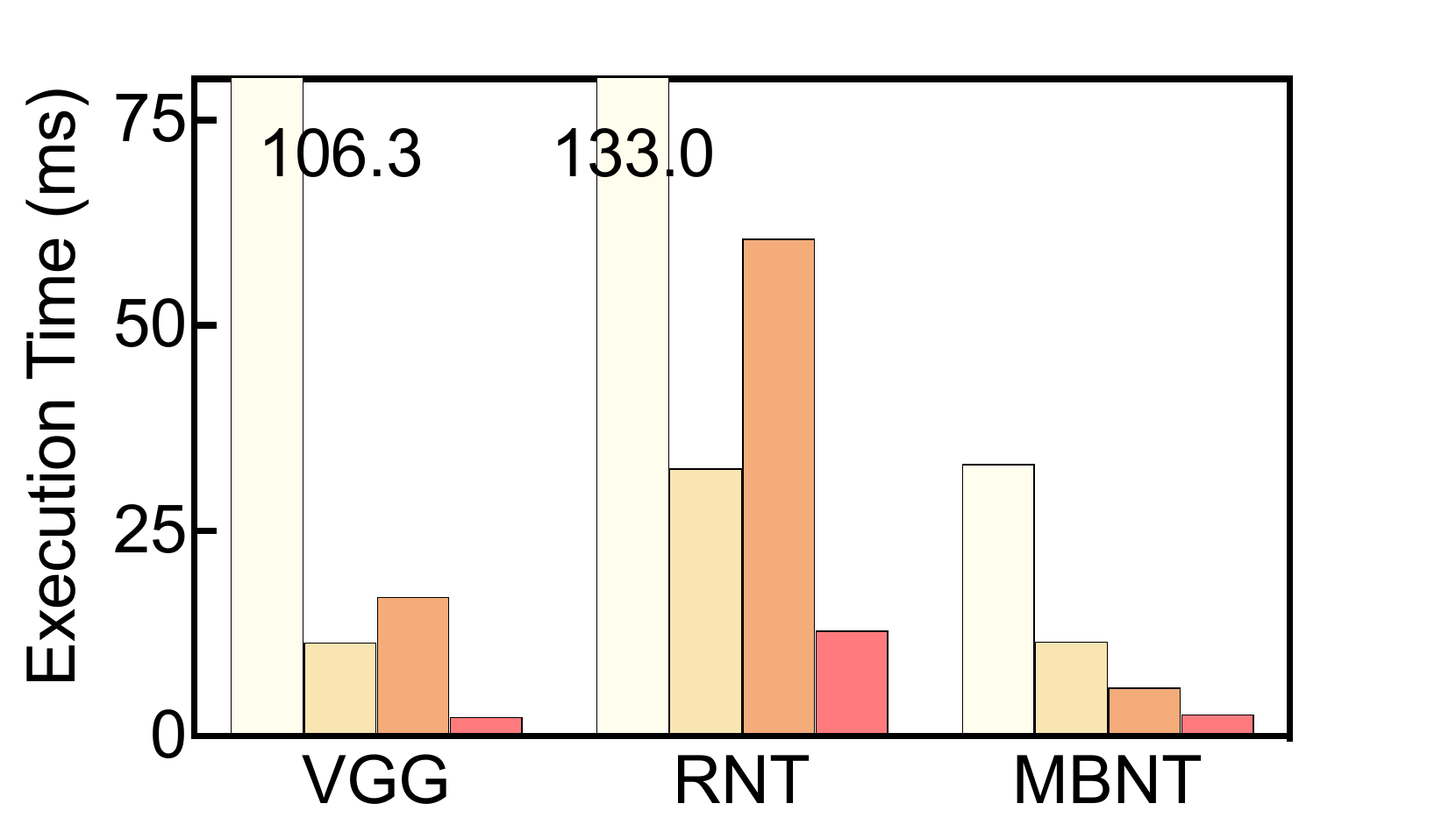}
        }
        \subfloat[ImageNet-GPU]{
            \includegraphics[width=0.25\textwidth]{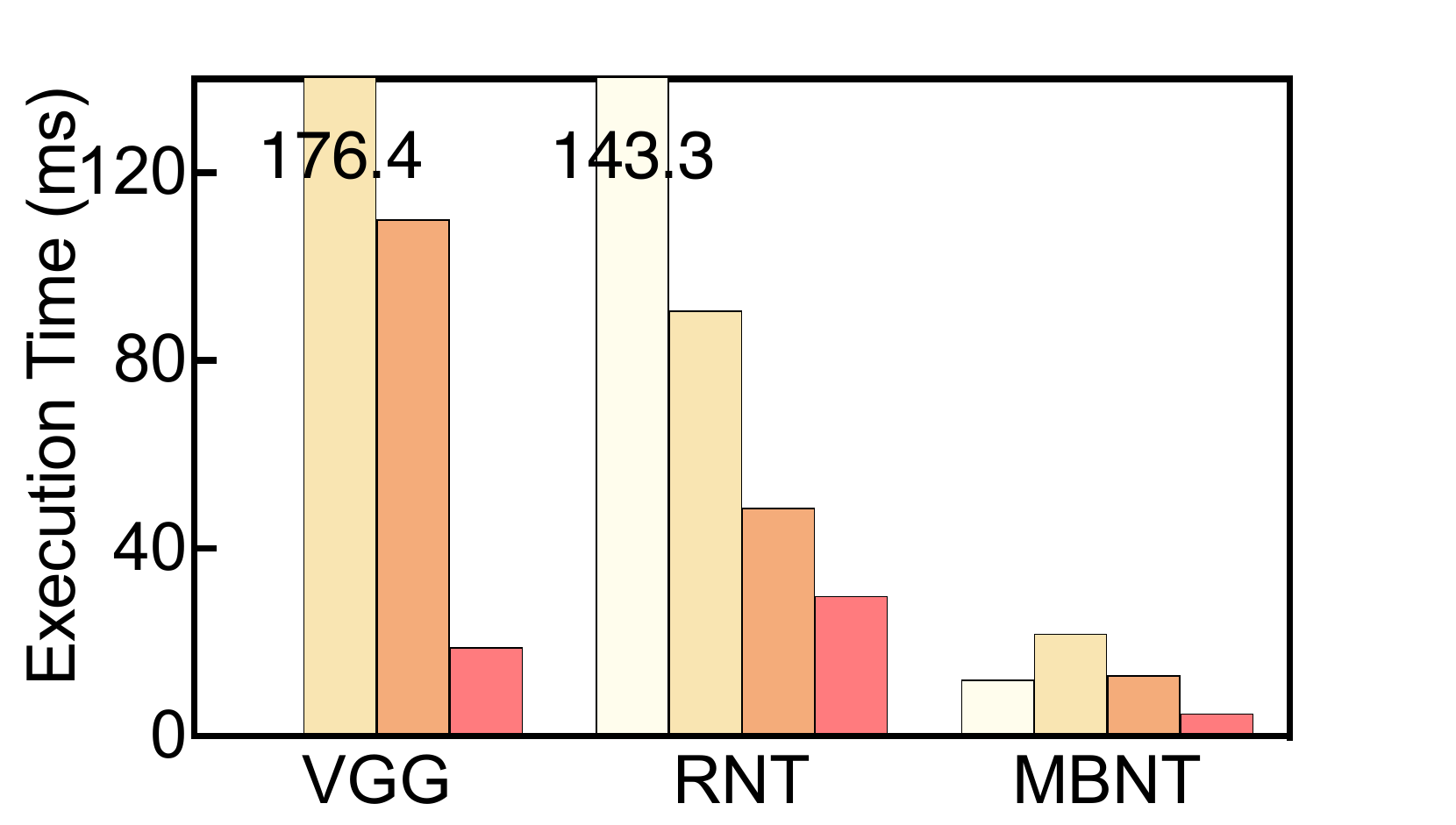}
        }
        \subfloat[CIFAR-10-GPU]{
            \includegraphics[width=0.25\textwidth]{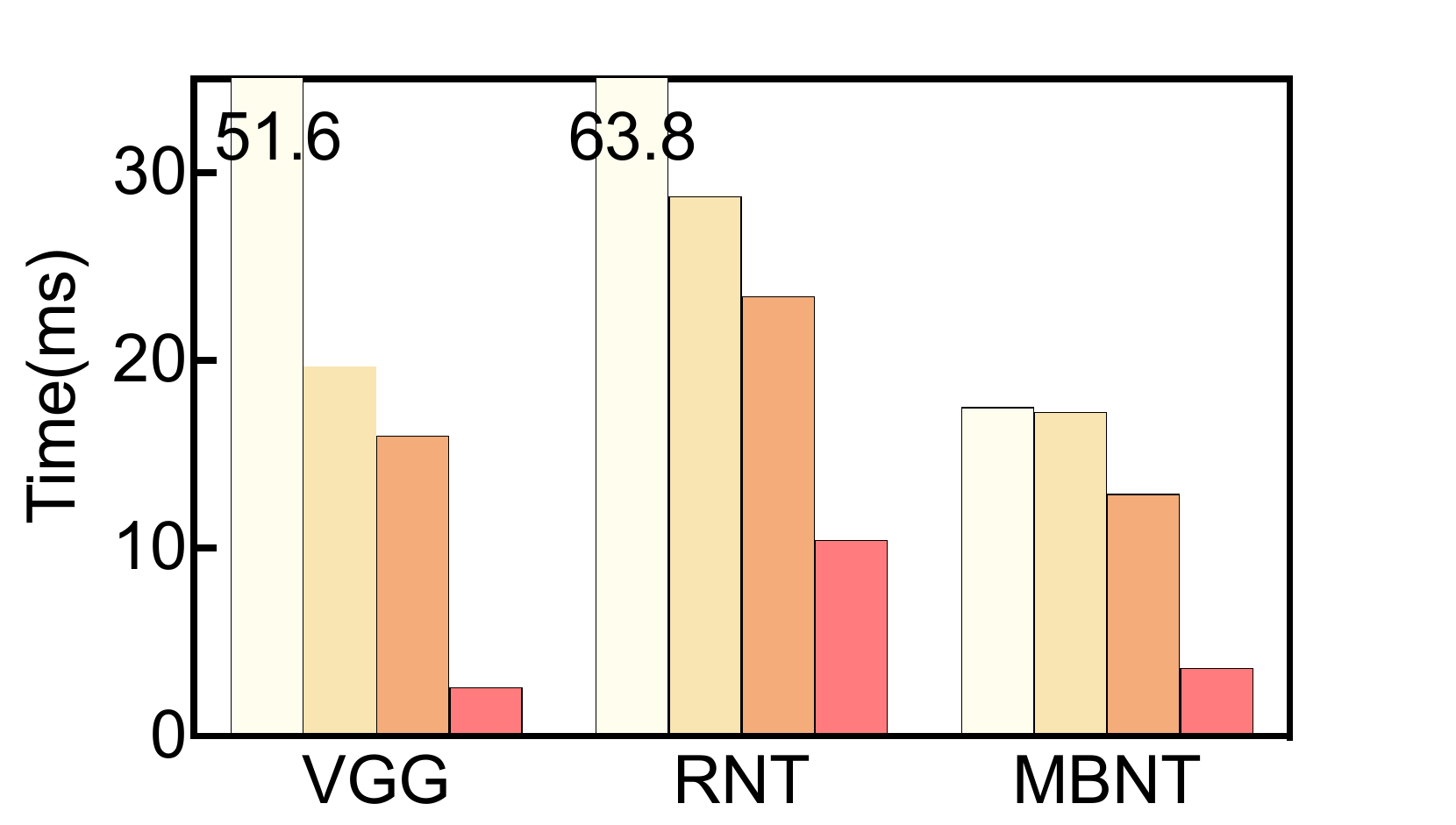}
        }
        \caption{Overall performance: x-axis: different trained DNN models; y-axis: average DNN inference execution time on a single input.}
    \label{fig:eva_overview_performance}
        \vspace{-1em}
\end{figure*}

\subsection{Methodology}
% Model profile

\noindent{\bf Evaluation Objective:} Our overall evaluation 
demonstrates that achieving real-time inference of large-scale DNNs on modern mobile devices is possible with PatDNN. Specifically, the evaluation has five objectives: 
(1) demonstrating that PatDNN outperforms existing state-of-the-art DNN frameworks 
%, e.g., achieving $1.6\times$ to $44.5\times$ 
without any accuracy compromise; 
(2) studying the performance effect of our key compiler optimizations 
%(filter kernel reorder, load redundancy elimination, and parameter tuning) 
and explaining the reasons for performance improvement; 
(3) further confirming the performance of PatDNN by comparing its pure GFLOPS with our optimized dense baseline; 
(4) showing that PatDNN performs similarly on different mobile platforms, i.e., PatDNN has a good portability; and 
(5) unveiling the impact of pattern count selections on both the accuracy and performance.       

\noindent{\bf DNNs and Datasets:} PatDNN is evaluated on three mainstream DNNs, VGG-16 (VGG), ResNet-50 (RNT), and Mobile-Net-V2 (MBNT). They are trained on two datasets, ImageNet and CIFAR-10. Table~\ref{tab:datasets} characterizes these trained DNNs.
%, including their layers counts, CONV layers counts, pattern counts, accuracy, and accuracy loss. 
Some information is omitted due to the space constraint, e.g., a uniform CONV pruning rate for VGG and RNT is $8\times$, and $4.4\times$, respectively (with uniform 3.6$\times$ connectivity pruning rate).
VGG has 13 CONV layers, and 5 of them have identical structures to others. Table~\ref{tab:eva_vgg_shortname} lists the filter shape ([{\tt \#output channel, \#input channel, kernel height, and kernel width}]) of these 9 unique layers and gives them a short name each.

\noindent{\bf Evaluation Platforms and Running Configurations:} Our experiments are conducted on a Samsung Galaxy S10 cell phone with the latest Qualcomm Snapdragon 855 mobile platform that consists of a Qualcomm Kryo 485 Octa-core CPU and a Qualcomm Adreno 640 GPU. 
%\textcolor{red}{To complete} 
Our portability tests are conducted on a Xiaomi POCOPHONE F1 phone with a Qualcomm Snapdragon 845 that consists of a Kryo 385 Octa-core CPU and an Adreno 630 GPU, and an Honor Magic 2 phone with a Kirin 980 that consists of an ARM Octa-core CPU and a Mali-G76 GPU.
All tests run 50 times on different input (images) with 8 threads on CPU, and all pipelines on GPU. Because multiple runs do not vary significantly, this section only reports the average time for readability. Because CONV layers are most time-consuming, accounting for more than 
%\textcolor{red}{$95\%$ ($90\%$ for VGG)} 
$95\%$ ($90\%$ for VGG)
of the total execution time, our evaluation focuses on the CONV layers. 
%In addition, it has been proved~\cite{} that 16-bit quantization on GPU results in optimized performance without compromising the prediction accuracy. So all reported GPU results apply 16-bit quantization. \textcolor{red}{Double check if this is correct...} 
%\textcolor{red}{Do we need mention GPU-16 here? Or show the GPU-32 results too?}
%\textcolor{red}{Without accuracy loss, Winograd, Uniform pruning, Pruning rate, Float-16 fair, non-structured no speedup at all}
All runs are tuned to their best configurations, e.g., Winograd optimization~\cite{lavin2016fast} is used for all dense runs, and 16-bit float point is used for all GPU runs.

\subsection{Overall Performance}

% \textcolor{red}{TODO: winograd is used for dense; no fully-connected}
% \textcolor{red}{TODO: effect of uniform pruning rate and for weight reduction.}

% Figure - Over performance compared with TFLite and MNN and TVM (CPU with gmacps)

Figure~\ref{fig:eva_overview_performance} shows the overall CPU and GPU performance of PatDNN compared to TFLite, TVM, MNN on all six trained DNNs. PatDNN outperforms all other frameworks for all cases. 
On CPU, PatDNN achieves $12.3\times$ to $44.5\times$ speedup over TFLite, $2.4\times$ to $5.1\times$ over TVM, and $1.9\times$ to $7.1\times$ over MNN, respectively. %On GPU, PatDNN achieves $2.5\times$ to $20\times$ speedup over TFLite, $2.8\times$ to $11.4\times$ speedup over TVM, and $1.6\times$ to $6.2\times$ speedup over MNN, respectively
On GPU, PatDNN achieves $2.5\times$ to $20\times$, $2.8\times$ to $11.4\times$, and $1.6\times$ to $6.2\times$ speedup over TFLite, TVM, and MNN, respectively\footnote{TFLite does not support executing VGG on ImageNet data set on GPU due to its too large memory footprint.}. For the largest DNN (VGG) and largest data set (ImageNet), PatDNN completes CONV layers on a single input  within 
%\textcolor{red}{18.9} 
18.9 ms on GPU.
%\textcolor{red}{need add "on which GPU and SOCs"?}. 
Even including the other rest layers (like FC), PatDNN can still meet the real-time requirement (usually 30 frames/sec, 
i.e., 33 ms/frame).

% Figure - Over performance compared with TFLite and MNN and TVM (CPU with gmacps)

%overall workload reduction

\begin{figure}[t]
    \centering
        \subfloat[CPU]{
            \includegraphics[width=0.485 \columnwidth]{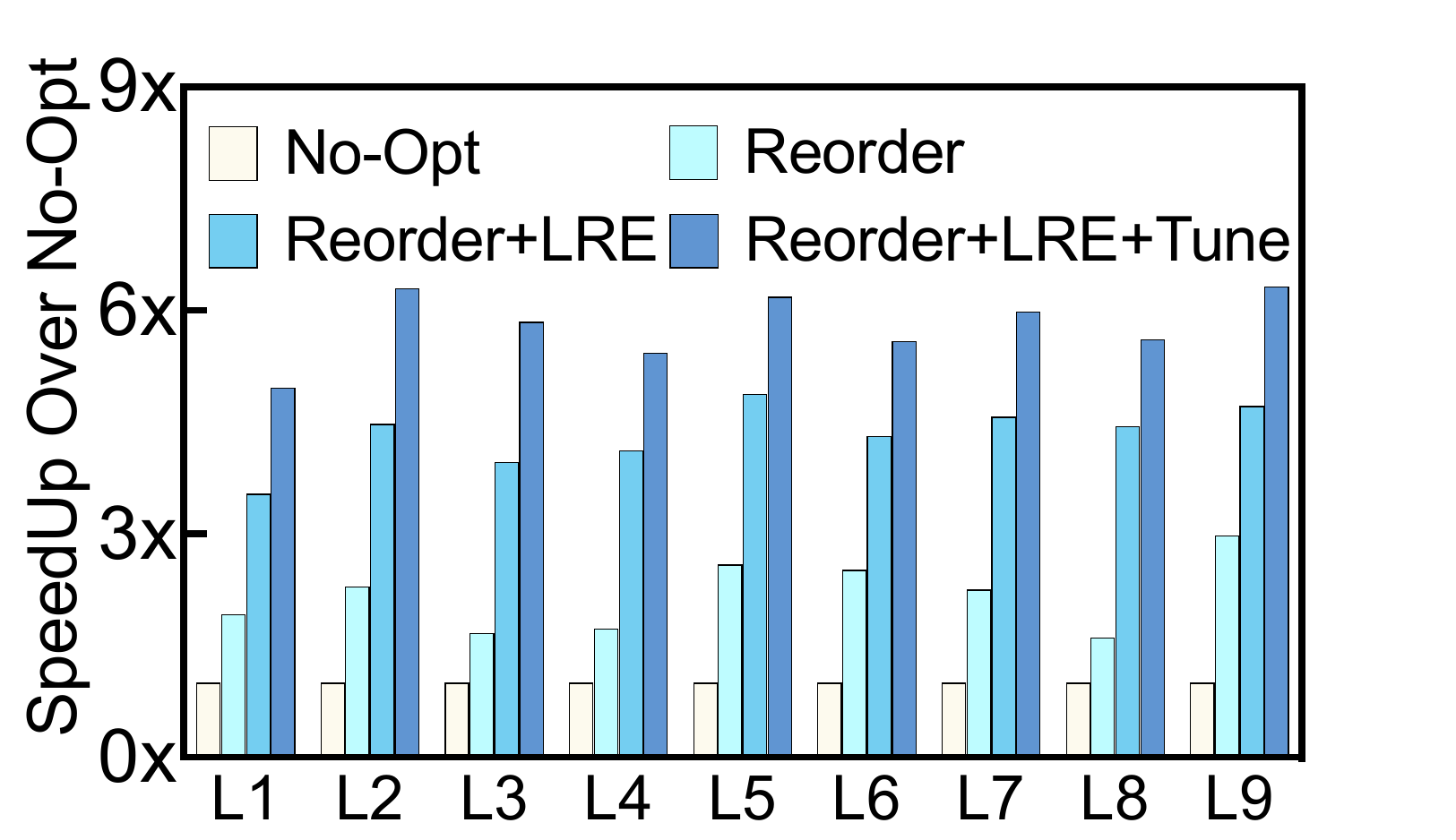}
        }
        \subfloat[GPU]{
            \includegraphics[width=0.5 \columnwidth]{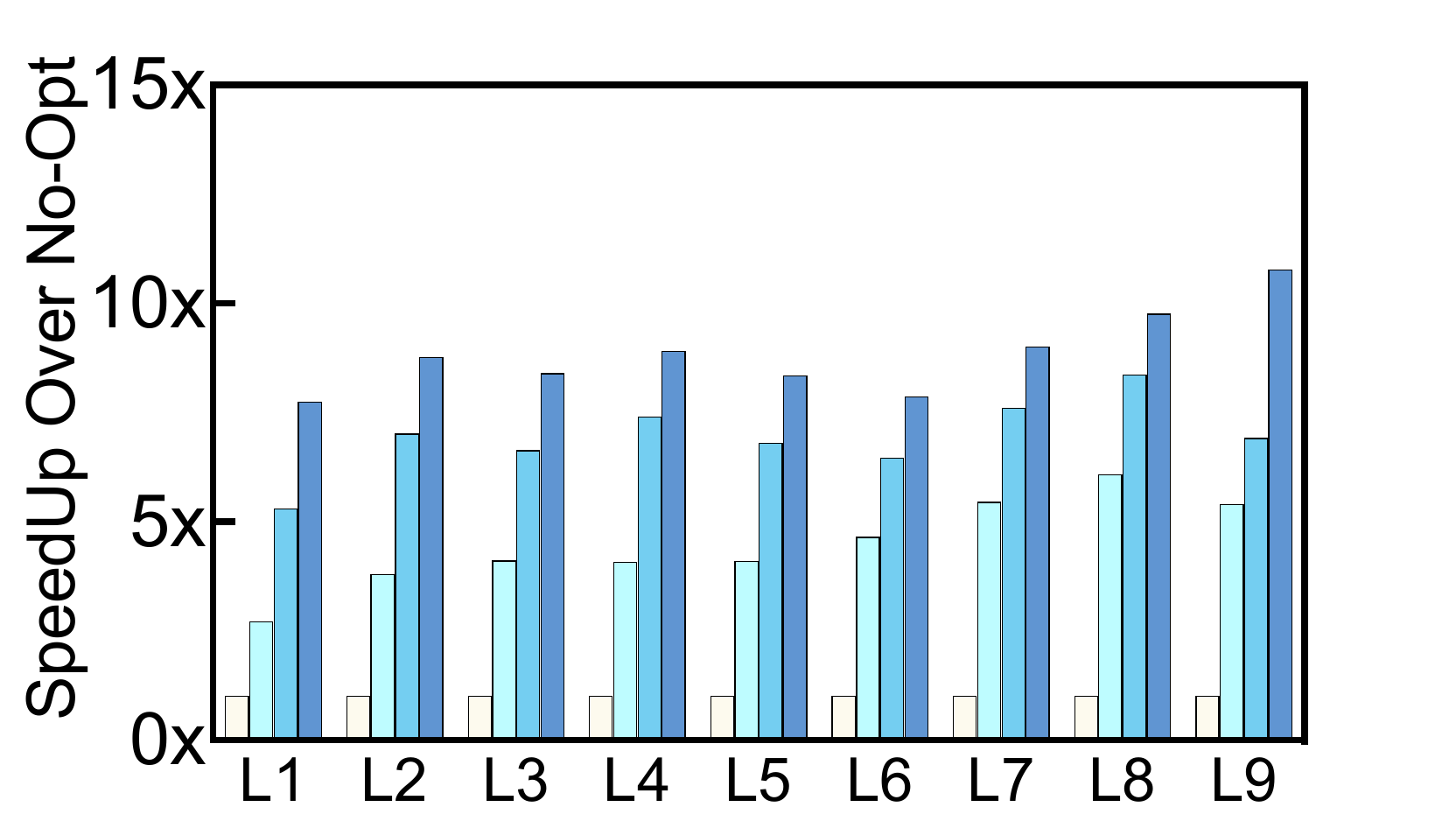}
        }
        \caption{Speedup of opt/no-opt on each unique CONV layer.}
    \label{fig:eva_reorder_rle_tuning_speedup}
        %\vspace{-1.5em}
\end{figure}

\begin{figure}[t]
    \centering
        \subfloat[Dense w/o Wino]{
            \includegraphics[width=0.267 \columnwidth]{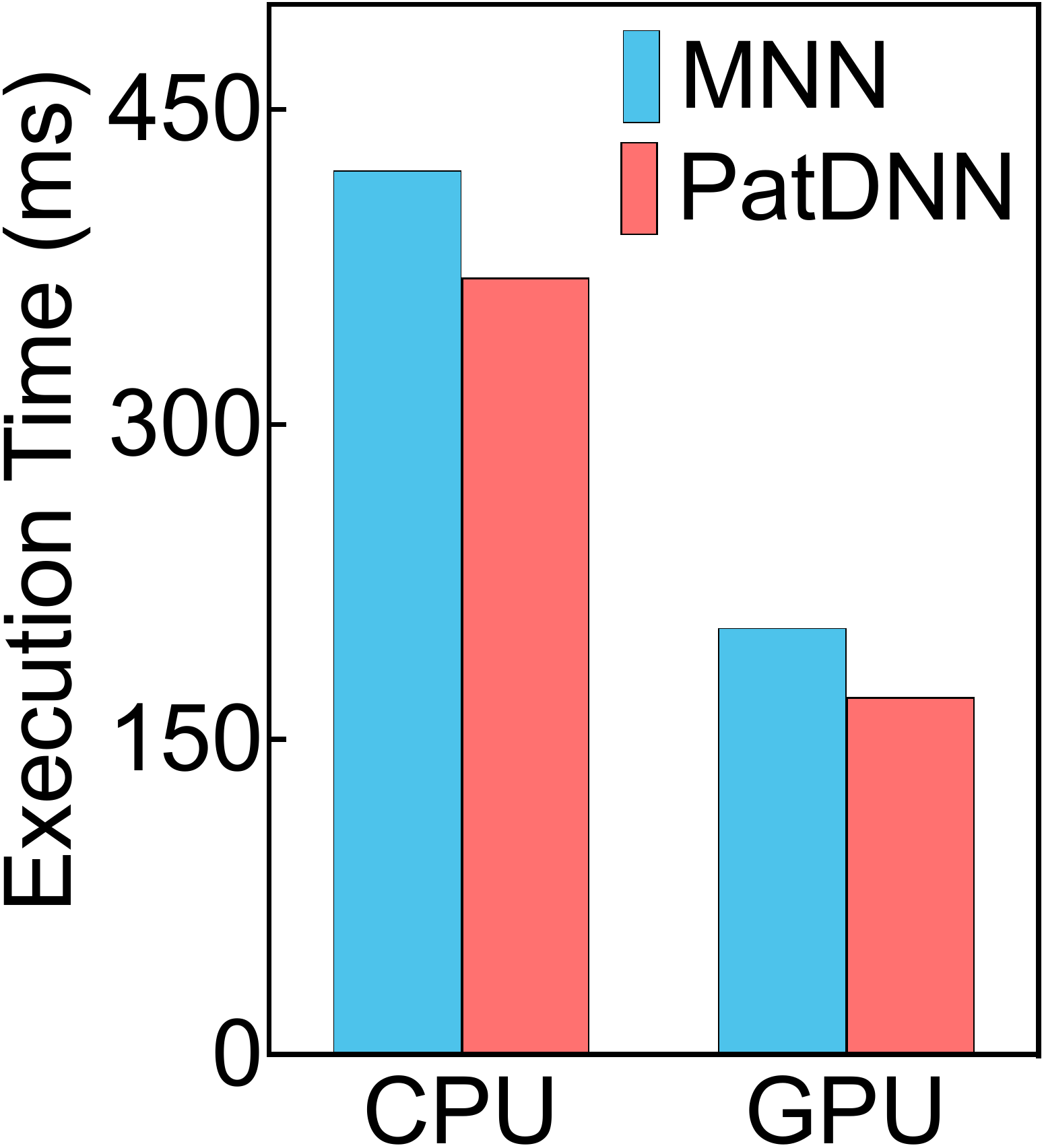}
        }
        \subfloat[Performance in GFLOPS: pattern vs dense]{
            \includegraphics[width=0.7 \columnwidth]{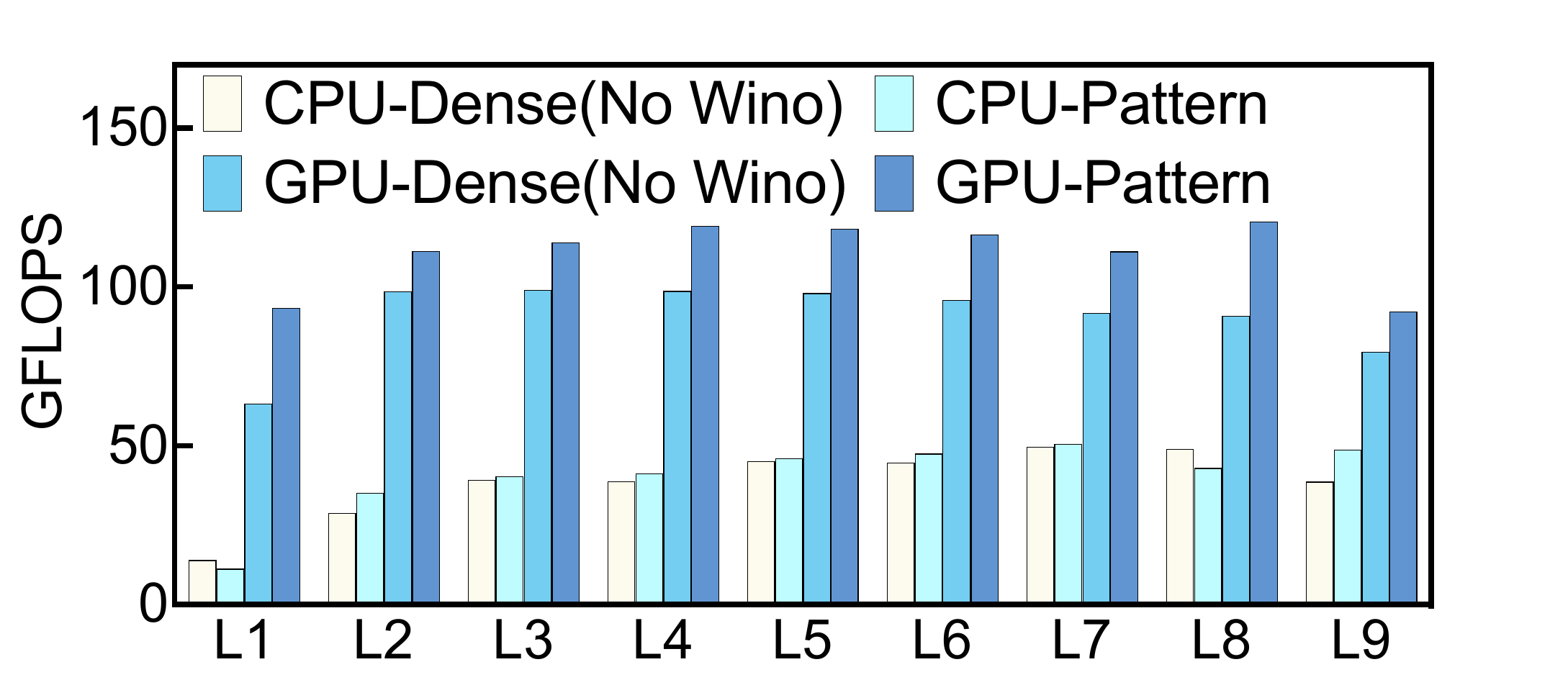}
        }
        \caption{GFLOPS performance study: PatDNN vs dense.}
    \label{fig:eva_dense_pattern_compare}
        %\vspace{-1em}
\end{figure}

%\blue{
PatDNN outperforms other frameworks because of two major reasons. First, its dense version is already $1.1\times$ to $1.6\times$ faster than TVM and MNN on mobile platforms because of some extra optimizations (as shown in Table~\ref{tab:dnn-frameworks}). Figure~\ref{fig:eva_dense_pattern_compare}(a) shows that PatDNN's dense version is faster than MNN on VGG, our largest DNN. Second, the pattern-based pruning reduces the overall computation by $3\times$ to $8\times$. Such computation reduction unfortunately cannot transfer to performance gains directly. %, i.e., without further optimization, PatDNN's compressed version performs almost the same to its dense version. 
We confirmed this by implementing an optimized sparse matrix version of PatDNN based on CSR~\cite{greathouse2016clsparse}, which shows almost the same speed to PatDNN's dense version. However, the subsequent compiler-level optimizations (filter kernel reorder, load redundancy elimination, auto-tuning, and compressed weight storage) successfully convert this computation reduction into real performance gains. We conduct a more detailed study on these optimizations in the next Section, and Figure~\ref{fig:eva_reorder_rle_tuning_speedup} shows a break-down of these optimizations’ contributions. Figures~\ref{fig:eva_reorder_rle_detail} to~\ref{fig:eva_csr_tight_storage} provide a detailed analysis of the underlying reasons. 
%}

\subsection{Optimization Evaluation}
% Four Computation Pattern Conv Kernel-Speed

%  Layer code fusion

%\subsection{Pattern-Related}

%\blue{Explain why PatDNN is faster than others frameworks}

This section studies the effect of our key compiler optimizations
%, filter kernel reorder (FKR), load redundancy elimination, and parameter tuning, 
and shows that our PatDNN's good performance mainly comes from these pattern-enabled optimizations. 
This part also compares the extra structure overhead between FKW and CSR.
Constrained by space, we only report the results of 
VGG, our most complex DNN, on the most widely accepted dataset (ImageNet). Experiments on other DNNs and datasets show the same trend. The rest parts also use VGG on ImageNet as a representative example.%\footnote{The same to the rest evaluations.}. 

%\blue{Also give the time numbers.}

Figure~\ref{fig:eva_reorder_rle_tuning_speedup} reports the speedup of the versions with optimizations over the version without any optimization on each unique CONV layer of VGG on CPU and GPU, respectively.  
On CPU, reorder brings $1.6\times$ to $3.0\times$ speedup, load redundancy eliminations bring additional $1.6\times$ to $2.8\times$ speedup, and parameter tuning brings additional $1.2\times$ to $1.9\times$ speedup. On GPU, these numbers are $2.7\times$ to $6.1\times$, $1.5\times$ to $3.3\times$ and $1.4\times$ to $3.8\times$. It is interesting that FKR brings more benefits on GPU than on CPU, because GPU's performance is more sensitive to the thread divergence and load balance due to its massive parallel nature. %\textcolor{red}{Any other interesting claims?}   
We next study why these optimizations work.
%Next part further studies the underlying reasons why these optimizations improve the performance.

\begin{figure}[t]
    \centering
        \subfloat[Filter length distribution before and after filter kernel reorder for L4]{
            \includegraphics[width=0.484 \columnwidth]{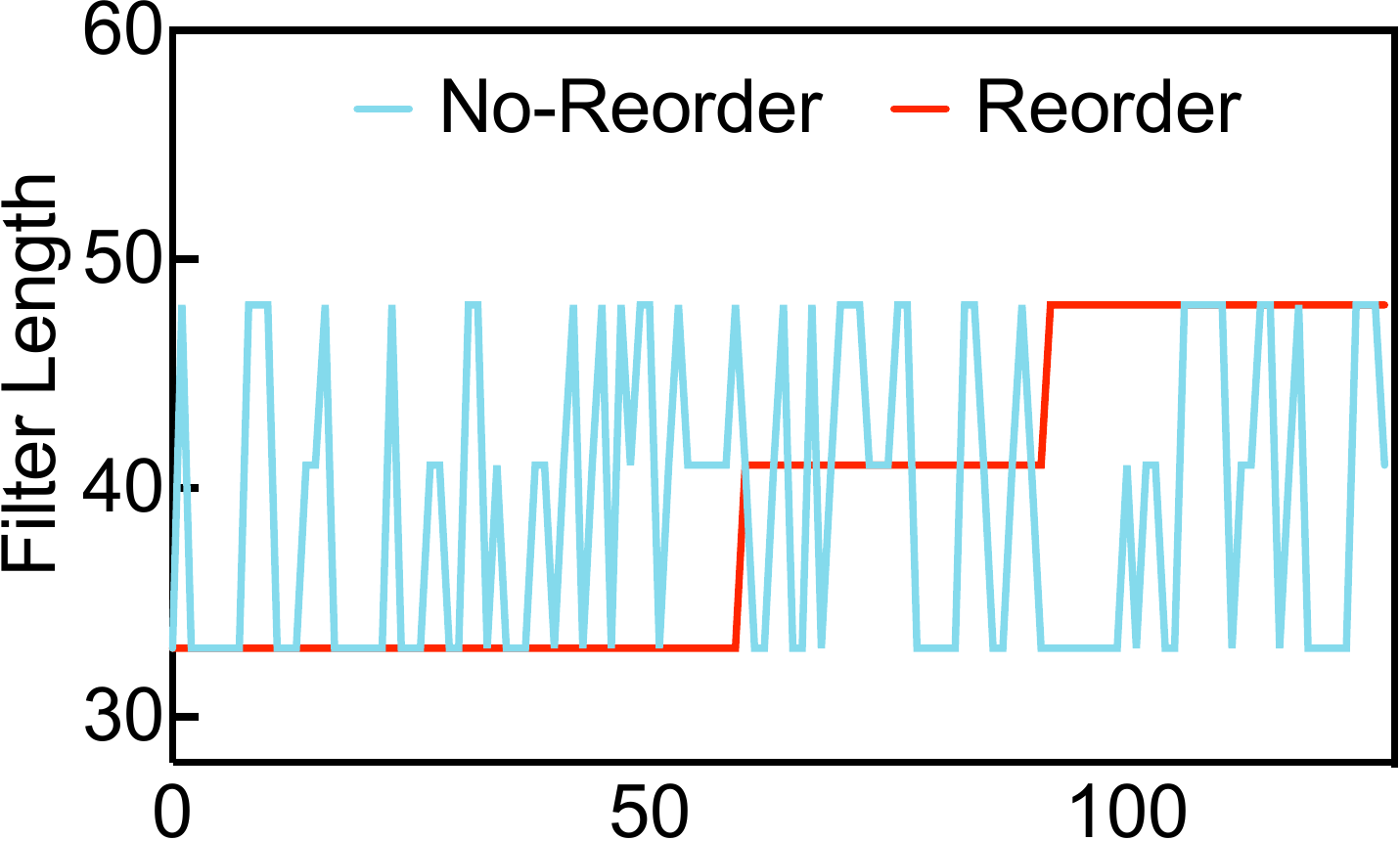}
        }
        \subfloat[Register load counts before and after elimination]{
            \includegraphics[width=0.516 \columnwidth]{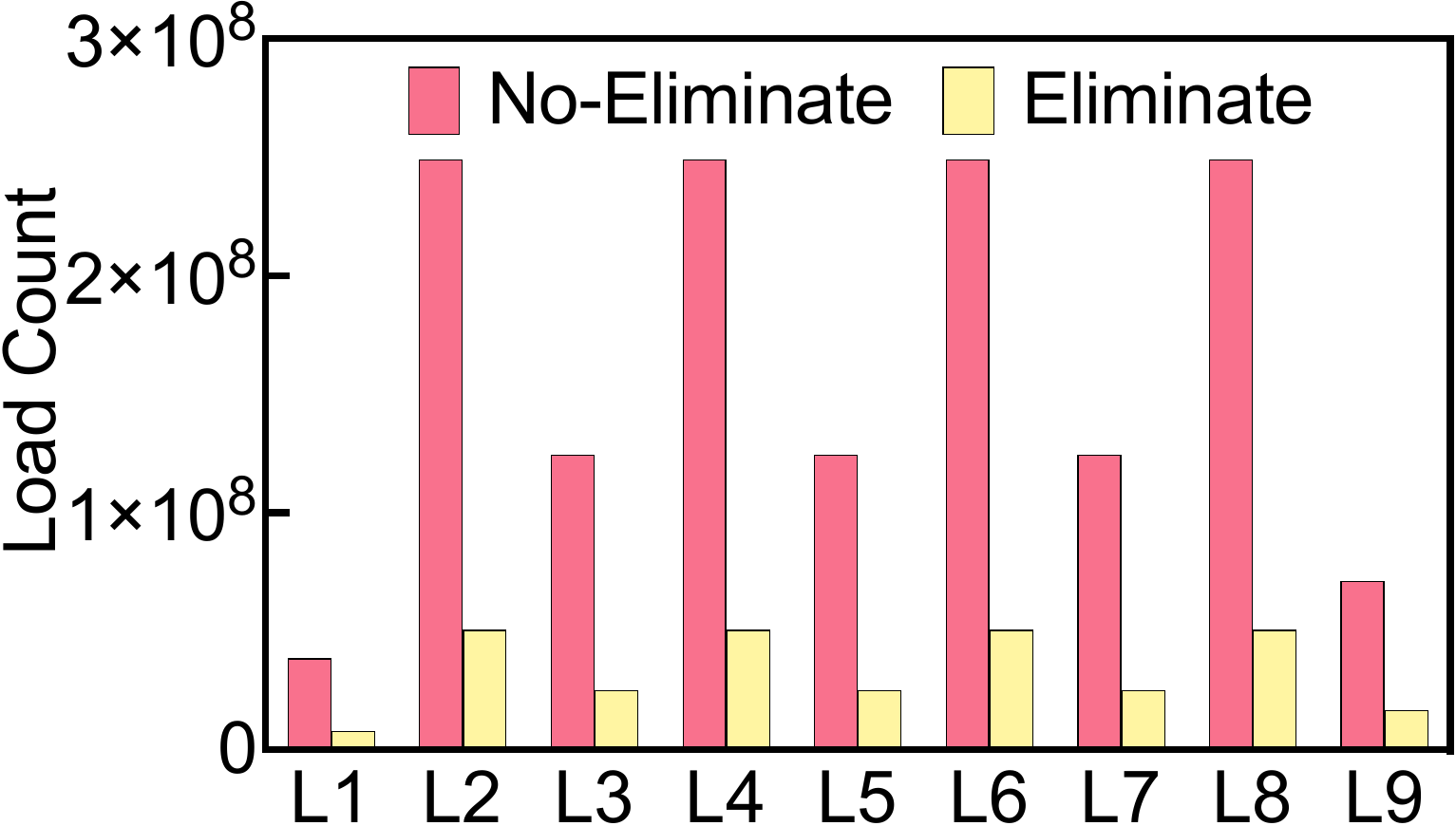}
        }
        \caption{Profiling result: reorder and redundancy elimination.}
    \label{fig:eva_reorder_rle_detail}
\end{figure}

\begin{figure}[t]
    \centering
        \subfloat[ImageNet]{\includegraphics[width=0.5 \columnwidth]{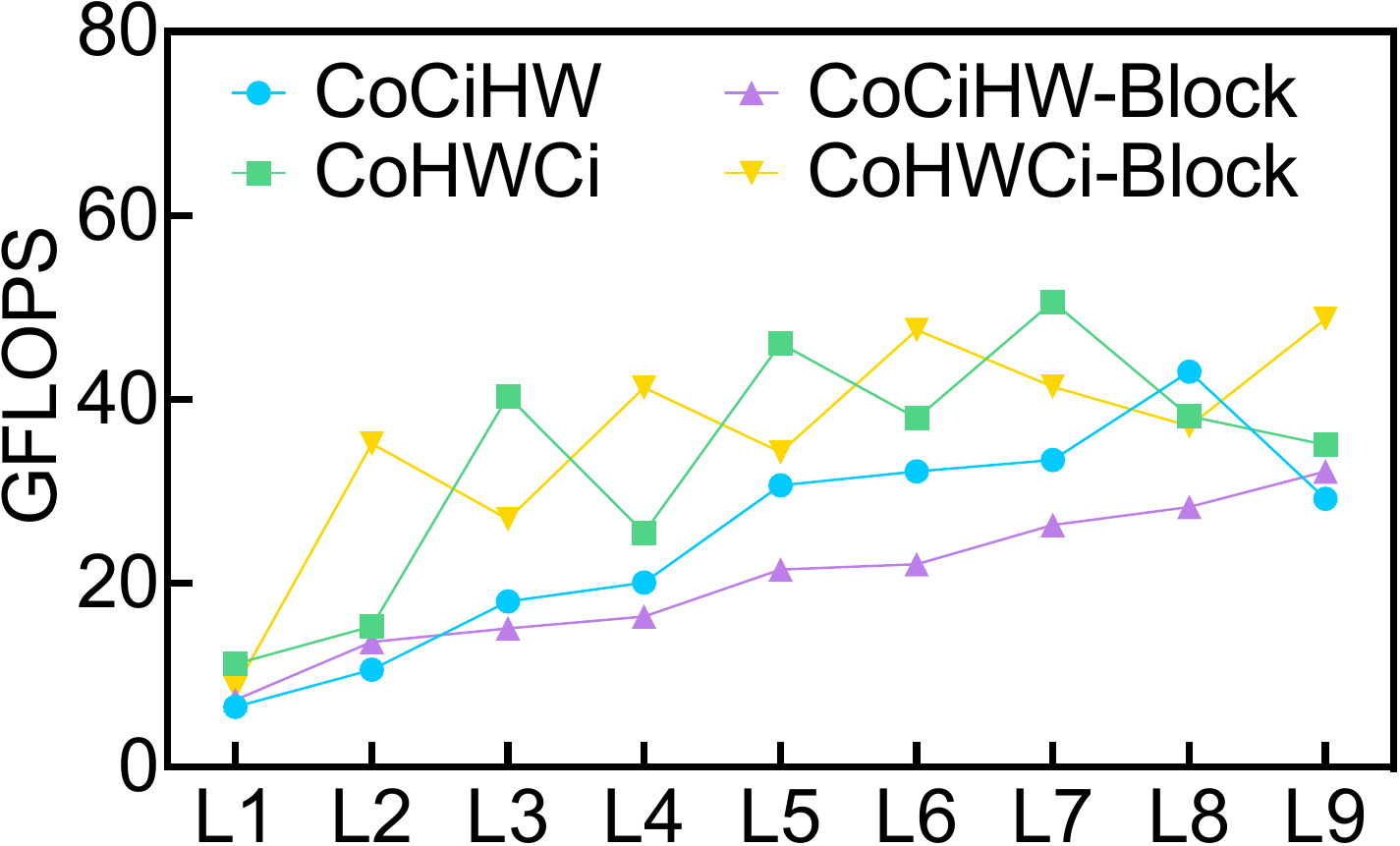}}
        \subfloat[CIFAR-10]{\includegraphics[width=0.5 \columnwidth]{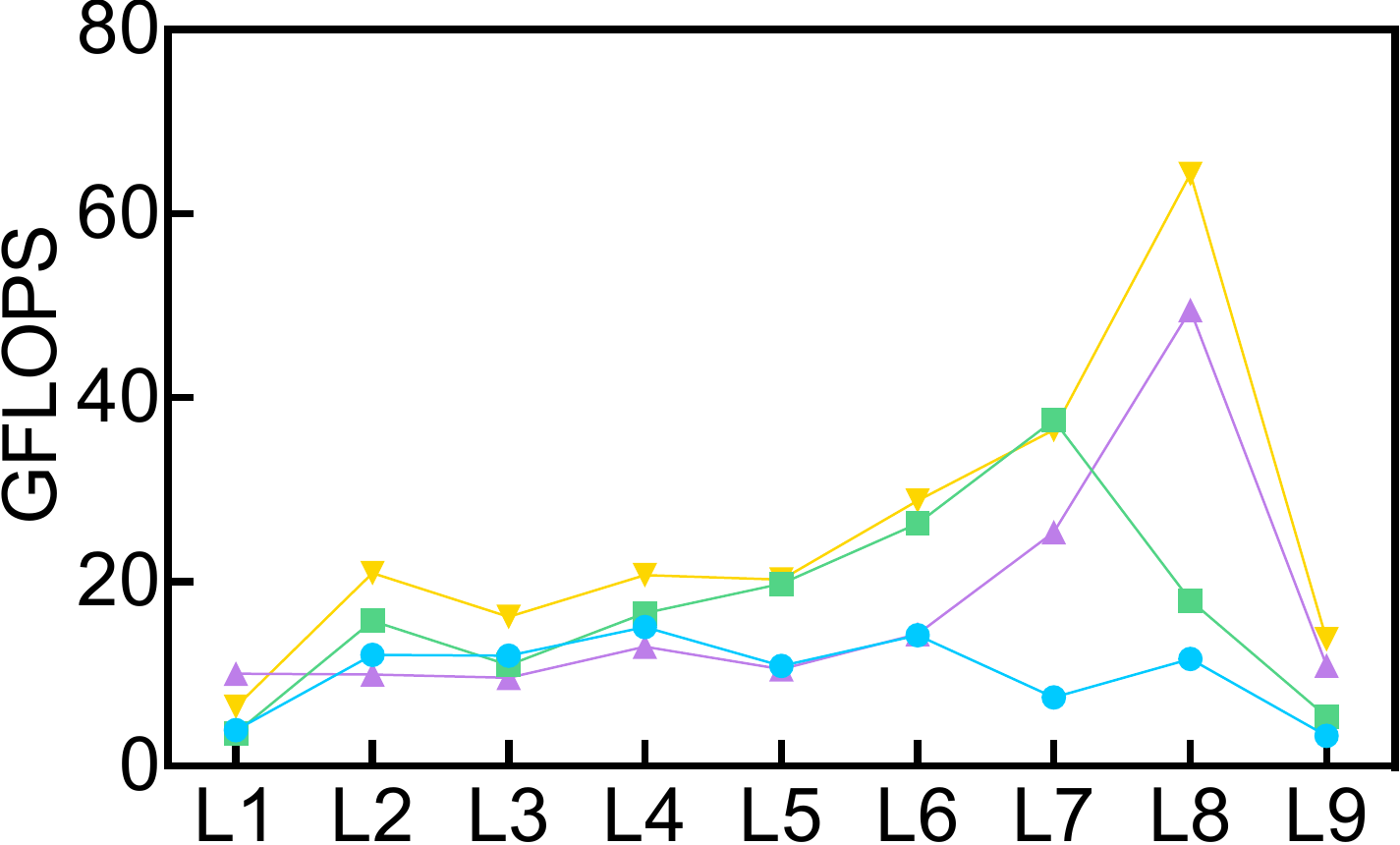}}
        \caption{Effect of different loop permutations and loop tiling.}
    \label{fig:eva_four_comp_tuning}
\end{figure}

\begin{figure}[t]
    \centering
    \includegraphics[width=0.45\textwidth]{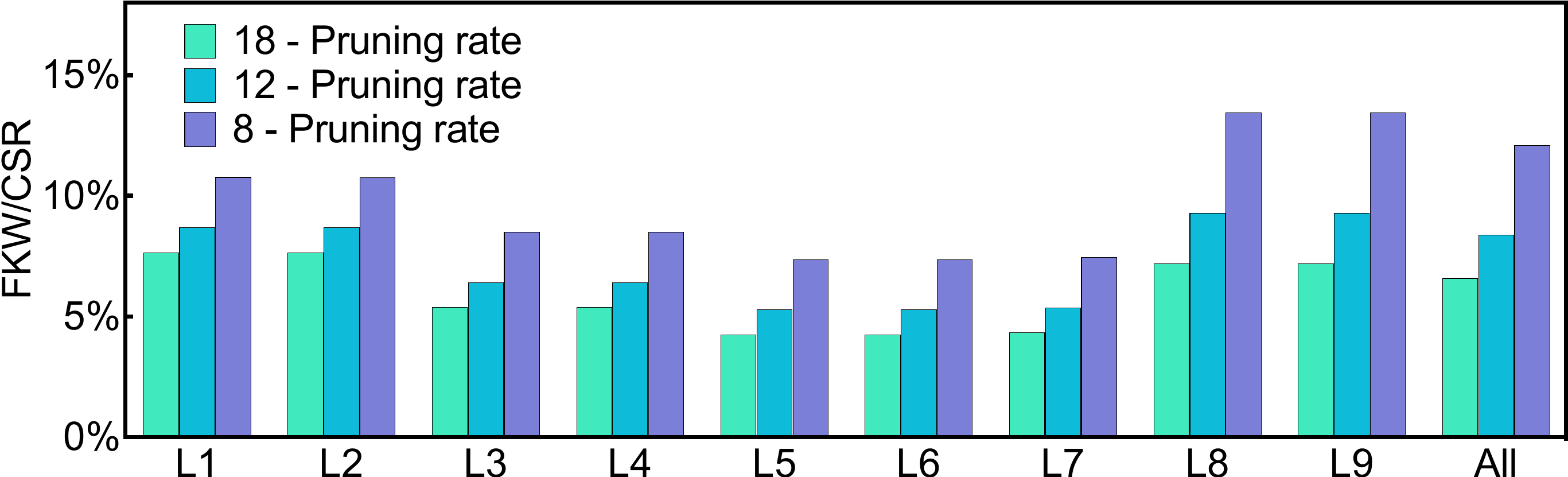}
    %\vspace{-1mm}
    \caption{Extra data structure overhead: FKW over CSR on unique VGG CONV layers with different pruning rates.}
    \label{fig:eva_csr_tight_storage}
    %\vspace{-2mm}
\end{figure}

\noindent{\bf Filter Kernel Reorder:}
%\subsubsection{Filter Kernel Reorder}
%performance before and after reorder; profiling on load balance; auto-tuning
Figure~\ref{fig:eva_reorder_rle_detail} (a) reports the filter length distribution of VGG L4 before and after FKR. Before reorder, the filters with varied lengths are distributed randomly, resulting in significant load imbalance if assigning them to different threads. After reorder, the filters are grouped into three groups, and the filters within each group have identical lengths. Each group could be executed by CPU threads simultaneously, or mapped to the same GPU thread block.

% \subsubsection{Connectivity Pruning}
% performance; workload reduction

\noindent{\bf Load Redundant Elimination:}
%\subsubsection{Load Redundant Elimination}
%performance; register load profiling
Figure~\ref{fig:eva_reorder_rle_detail} (b) reports the register load counts before and after LRE for each unique CONV of VGG. It shows that our register LRE can significantly reduce the number of register loads. 
Note that even if register load has lower
latency than cache or memory load, the memory/cache
performance has nevertheless been aggressively
optimized by conventional tiling. 
Thus, the significant performance gains
must have been achieved with the reduced
number of register loads. 
%That is why the performance gains are so obvious even the register load has lower latency than the cache or memory load, specifically considering the memory/cache performance has already been aggressively optimized by conventional tiling.  

\noindent{\bf Auto-tuning:} 
%\subsubsection{Auto-tuning} 
Figure~\ref{fig:eva_four_comp_tuning} reports 
the CPU performance (in GFLOPS) of each unique VGG CONV layer with varied loop permutations, and with or w/o 
%\textcolor{red}{Both of them with the best tilling size} 
blocking on ImageNet and CIFAR-10, respectively. It shows that different inputs and layers may require different configurations. 
Proper tuning will bring significant benefits. Constrained by space, we omit the GPU results and tuning results about GPU data placement.

\noindent{\bf Compressed Weight Storage:}
%\subsubsection{Compressed Weight Storage}
%model size in our format and csr
Figure~\ref{fig:eva_csr_tight_storage} shows the extra data structure overhead 
(i.e., the size of data structures other than weights) of FKW over CSR on each unique VGG CONV layer with three kinds of pruning rates, $18\times$, $12\times$, and $8\times$ respectively. For each one, FKW saves 93.4\%, 91.6\%, and 87.9\% extra data structure overhead over CSR in total, resulting in 46.7\%, 45.8\%, and 43.9\% overall storage space saving.

\subsection{PatDNN Performance Analysis in GFLOPS}
%compare the GFLOPS with optimized dense method.
To further analyze the performance of PatDNN, this part compares its pure GFLOPS with our dense implementation. To conduct an apple-to-apple comparison, we turn off the Winograd optimization that transforms the convolution operation to matrix-multiplication for a trade-off between the computation reduction and operation conversion overhead. 
Figure~\ref{fig:eva_dense_pattern_compare} (a) shows that our dense version can serve as an optimized baseline, because it is even faster than MNN.

Figure~\ref{fig:eva_dense_pattern_compare} (b) shows that our pattern-based (sparse) PatDNN achieves comparable GFLOPS to our optimized dense baseline on CPU, and outperforms it on GPU. It implies that the memory performance of PatDNN is comparable to the dense baseline on CPU and even better than it on GPU. This benefits from our model compression and memory load (and register load) reductions. Without pattern-based pruning, the input, output, and DNN model compete for the limited memory/cache resource; after pruning, only the input and output compete for it. PatDNN also reduces the overall computation; thus, it significantly outperforms all other mobile frameworks. %\textcolor{red}{traditional weight pruning results in no performance gain at all}
We cannot achieve this performance without our pattern-based design, and our other sparse implementation with conventional sparse matrix optimizations can only get either comparable or even slower speed than other mobile frameworks.

\begin{figure}[t]
    \centering
        \subfloat[Kirin 980]{
            \includegraphics[width=0.485 \columnwidth]{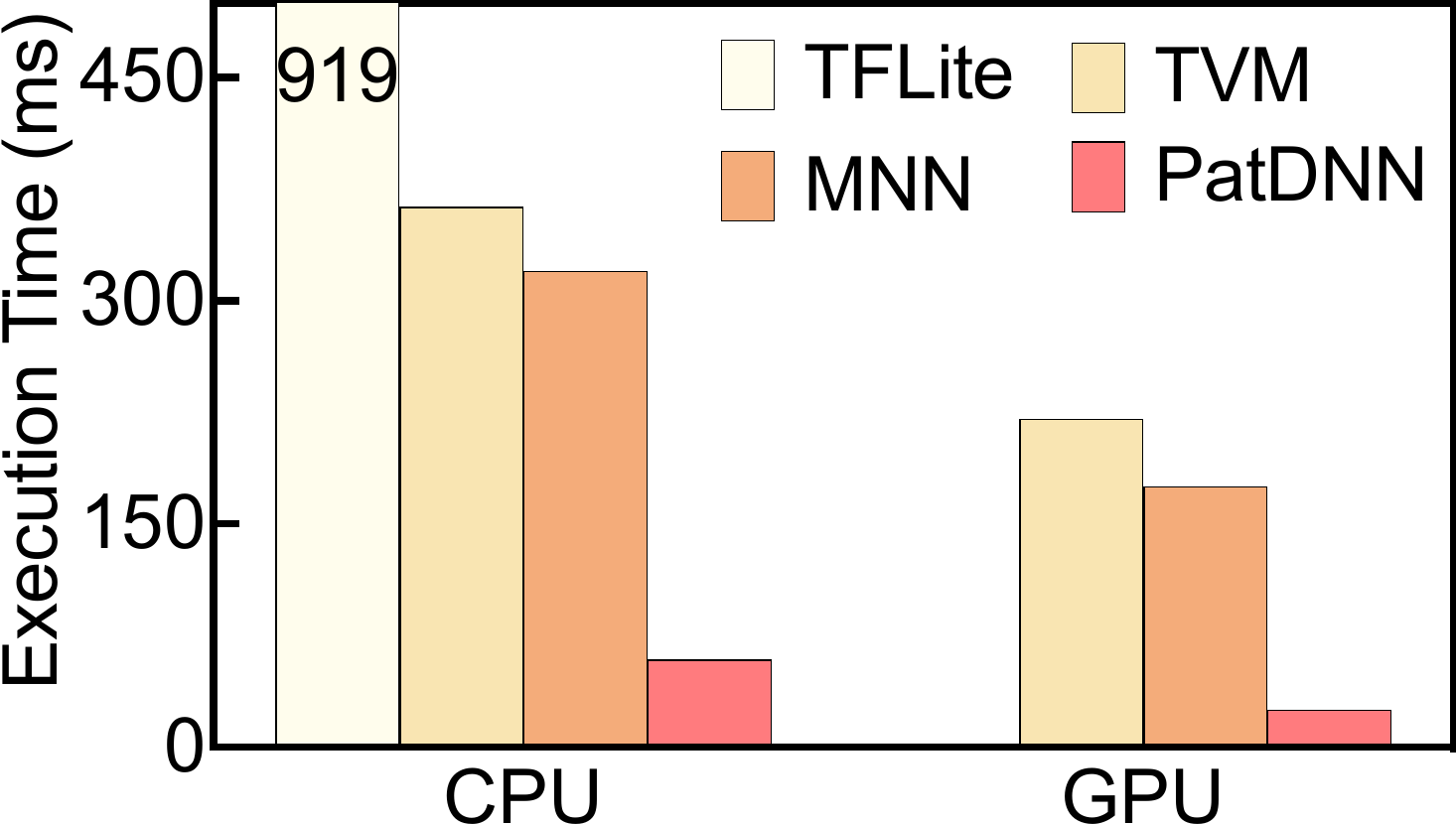}
        }
        \subfloat[Snapdragon 845]{
            \includegraphics[width=0.485 \columnwidth]{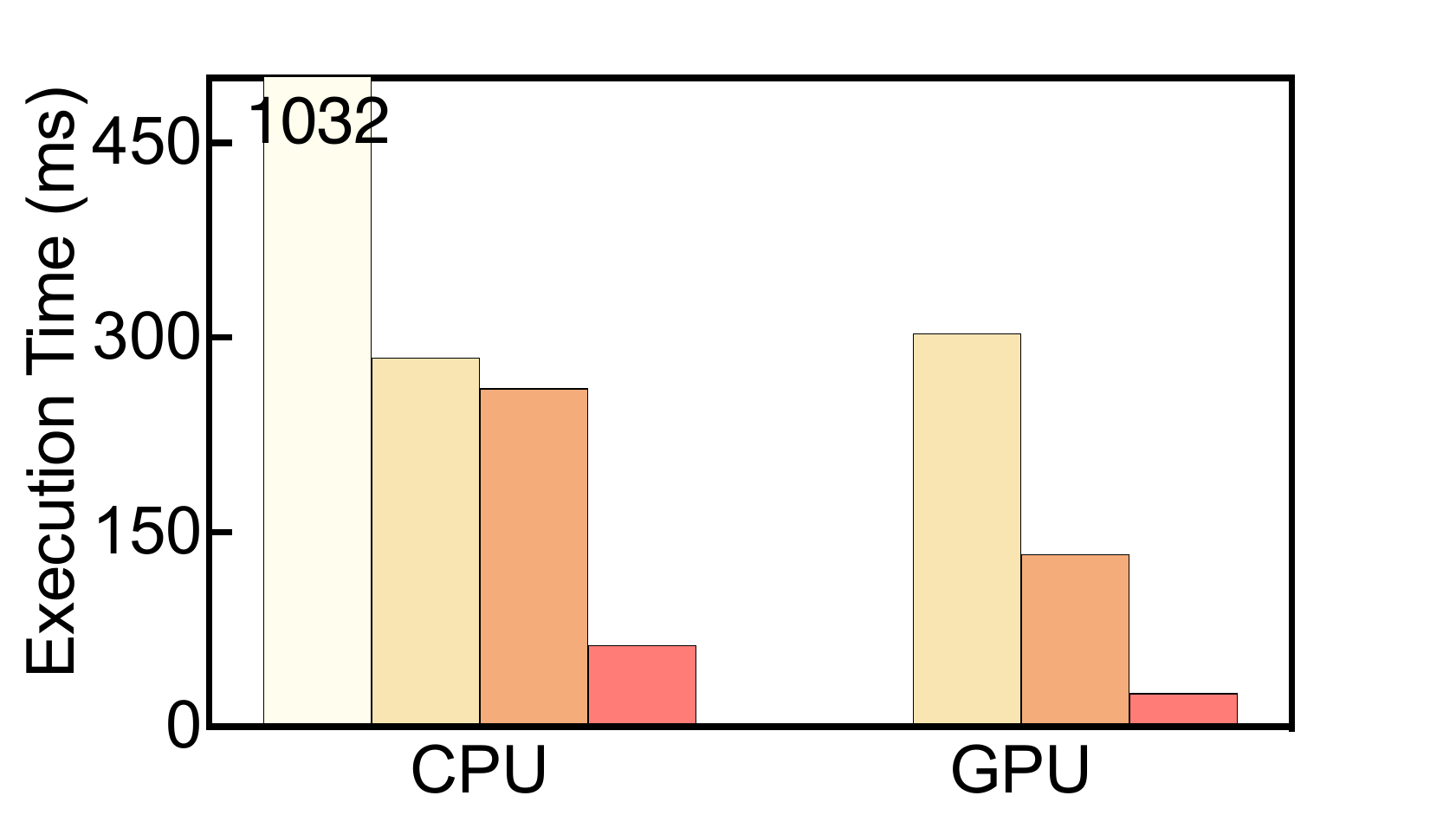}
        }
        \caption{Portability study: performance on two other platforms.}
    \label{fig:eva_portability}
\end{figure}

\begin{table}[t]
\vspace{-5mm}
\caption{Pattern counts impact (with 3.6$\times$ connectivity pruning): accuracy loss and exe time for VGG.}
\label{tab:table_patterns_accu_speed}
\centering
\includegraphics[width=1\linewidth]{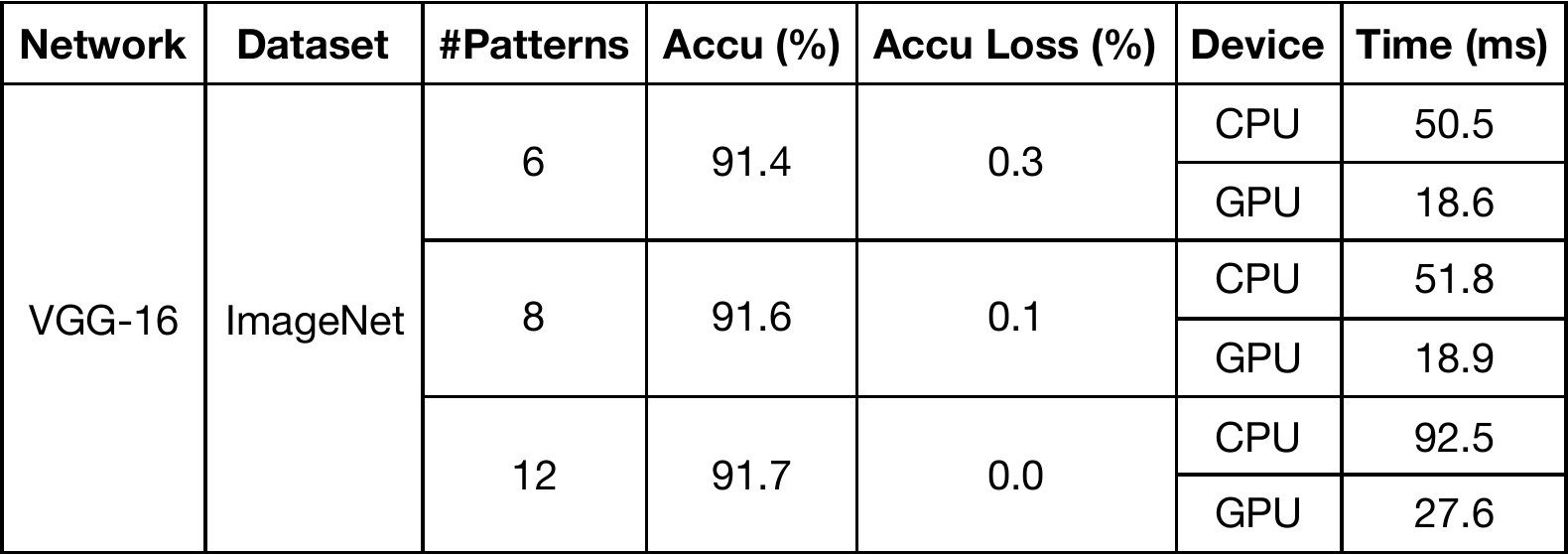}
\vspace{5mm}
\end{table}

\subsection{Portability Study}

PatDNN is also evaluated on two other platforms to confirm its portability. Figure~\ref{fig:eva_portability} shows the result. On these platforms, PatDNN also outperforms other frameworks. Particularly, other frameworks run much slower on Magic 2 than on Snapdragon 855; however, PatDNN performs more stably.
%\blue{
This is because our pattern-based pruning leads to fewer computations and fewer memory accesses thus reducing the memory bandwidth pressure.
%}

%\subsection{Energy Efficiency Comparison}

\subsection{Impact of Pattern Counts}
% framework profile

Table~\ref{tab:table_patterns_accu_speed} reports the impact of the pattern count selection on both the accuracy and execution time, under 3.6$\times$ uniform connectivity pruning rate. As increasing pattern counts, the accuracy increases slightly, however, the performance drops quickly. Our evaluation selects 8 patterns that result in ideal performance with a negligible accuracy loss.    

%% file: tex/conclusion.tex
\section{Discussion}

\noindent{\bf Generality:} The techniques proposed in PatDNN are general enough to be applied to other platforms. Compared to laptops or servers, mobile platforms are more resource-constrained, making it is more challenging to achieve real-time execution.  However, the need for real-time DNN execution
is crucial due to many important mobile applications.
%there are many important applications on mobile platforms that rely on real-time DNN execution. 
In fact, in addition to the mobile platforms in our paper, we also tested PatDNN on the latest Raspberry Pi 4 platform. It shows a similar speedup over other frameworks like TVM. We believe that it is a promising research direction to improve PatDNN's portability by incorporating it with TVM that emphasizes the DNN execution on varied computing devices.

\noindent{\bf Dense vs. Sparse DNNs:}
%\blue{
General end-to-end DNN inference acceleration frameworks like TFLite, TVM, and MNN do not support sparse DNN execution. If we simply add sparse DNN support with random pruning and general compression storage (like CSR) to these frameworks, it is expected that their speed cannot be improved significantly as shown in 
the results of PatDNN's CSR implementation. Although there is potential to improve the performance with coarse-grained structured pruning (that prunes whole filters/channels), the accuracy will be obviously degraded as we discussed before. From this perspective, PatDNN opens a new door to accelerate DNN execution with a compression/compiler-optimization co-design. With such co-design, sparse (or compressed) DNN execution becomes a more promising solution in resource-constraint environments than dense DNN.
%}

\section{Conclusion}\label{sec:conclusion}

This paper presents PatDNN, an end-to-end framework to achieve real-time DNN execution on mobile devices. PatDNN consists of two stages, a pattern-based pruning stage based on extended ADMM solution framework, and an optimized execution code generation stage including a high-level, fine-grained DNN layerwise representation and a set of archi-tecture-aware optimizations.
%, filter kernel reordering, compressed weight storage, register load redundancy eliminations, and parameter auto-tuning. 
This design allows PatDNN to benefit from both high accuracy and hardware efficiency. Our evaluation results demonstrate that PatDNN outperforms other state-of-the-art end-to-end DNN execution frameworks
%, such as TensorFlow Lite, TVM, and Alibaba Mobile Neural Network 
with up to $44.5\times$ speedup and no accuracy compromise, and achieves real-time execution of large-scale DNNs on mobile devices.  

%% file: tex/ack.tex
\section*{Acknowledgements}

The authors would like to thank the anonymous reviewers for their valuable and thorough comments. The authors are especially grateful to the shepherd
Yufei Ding for her extensive feedback and constructive suggestions that help improve this paper substantially. This work was
supported in part by the NSF awards CNS-1739748, CCF-1937500, CCF-1919117, CCF-1901378, and CCF-1919289.